\documentclass[12pt]{article}

\usepackage{wrapfig}
\usepackage{titlesec}
\usepackage{xcolor}
\usepackage{mathpazo}
\usepackage{ebgaramond}
\usepackage{charter}
\usepackage{bbold}
\usepackage{booktabs}
\usepackage{array}
\usepackage{multirow}
\usepackage{pifont}
\usepackage{xcolor}
\usepackage{pdflscape}
\usepackage{adjustbox}
\usepackage{subcaption}
\usepackage{colortbl} 
\usepackage{longtable} 
\usepackage{changepage}

\usepackage{amsthm}

\definecolor{customgray}{gray}{10.5}  
\newcommand{\mycomment}[1]{}
\titleformat{\section}
{\normalfont\Large\bfseries\color{customgray}} 
{\thesection}{1em}{}

\titleformat{\subsection}
{\normalfont\large\bfseries\color{gray}} 
{\thesubsection}{1em}{}
\titleformat{\subsubsection}
{\color{gray}\normalsize\bfseries} 
{\thesubsubsection}{1em}{}
\titlespacing*{\section}{0pt}{1.5ex plus 1ex minus .2ex}{1.5ex plus .2ex}
\titlespacing*{\subsection}{0pt}{1.25ex plus 1ex minus .2ex}{1.25ex plus .2ex}
\newtheorem{theorem}{Theorem}

\usepackage{times}
\usepackage{tabularray}
\usepackage{heuristica}
\usepackage{wrapfig}
\usepackage{multirow}
\usepackage{booktabs}
\usepackage{algorithm}
\usepackage{algpseudocode}

\usepackage[hmargin=1cm]{geometry}

\usepackage{tikz}
\usetikzlibrary{backgrounds}
\usetikzlibrary{patterns}
\usetikzlibrary{bayesnet}
\usetikzlibrary{arrows}
\usepackage{color}

\usetikzlibrary{backgrounds}

\usepackage{array}

\usepackage{amsmath,amsfonts}
\usepackage{amssymb} 
\usepackage{mathtools}

\usepackage{listings}
\usepackage[most]{tcolorbox}

\definecolor{usercolor}{RGB}{255, 108, 108} 
\definecolor{assistantcolor}{RGB}{255, 189, 69} 

\newtcolorbox{userbox}{
    colback=usercolor!5!white,
    colframe=usercolor!75!black,
    width=\linewidth-1cm,
    left=0.5cm,
    right=0.5cm,
    boxrule=0.4mm,
    arc=2mm,
    fonttitle=\bfseries,
    title=Prompt,
    breakable
}

\newtcolorbox{assistantbox}{
    colback=assistantcolor!5!white,
    colframe=assistantcolor!75!black,
    width=\linewidth-1cm,
    right=0.5cm,
    left=0.5cm,
    boxrule=0.4mm,
    arc=2mm,
    fonttitle=\bfseries,
    title=Assistant,
    breakable
}

\lstdefinestyle{pythonstyle}{
    language=Python,
    basicstyle=\ttfamily\small\color{black},           
    keywordstyle=\color{teal}\bfseries,                 
    stringstyle=\color{brown},                          
    commentstyle=\color{gray},                          
    numberstyle=\tiny\color{gray},                      
    frame=single,                                       
    backgroundcolor=\color{gray!5},                     
    breaklines=true,                                    
    showstringspaces=false,                             
    numbers=left,                                       
    xleftmargin=1em,                                    
    rulecolor=\color{black},  
    morecomment=[s][\color{olive!80!black!80}\slshape]{"""}{"""},  
}

\lstdefinestyle{bashstyle}{
    language=bash,
    basicstyle=\ttfamily\small,
    frame=single,
    backgroundcolor=\color{gray!10},
    breaklines=true,
    showstringspaces=false
}

\usepackage{amsmath}
\usepackage{amssymb}
\usepackage{mathtools}
\usepackage{amsthm}
\theoremstyle{plain}

\newtheorem{lemma}[theorem]{Lemma}

\theoremstyle{definition}

\theoremstyle{remark}

\definecolor{bronze}{RGB}{205, 127, 50}
\definecolor{silver}{RGB}{192, 192, 192}
\definecolor{gold}{RGB}{255, 215, 0}
\usepackage{graphicx}   
\usepackage{amsmath}    
\usepackage{amssymb}    
\usepackage{fancyhdr}   
\usepackage{geometry}   
\usepackage{titlesec}   
\usepackage{xcolor}     
\usepackage{caption}  
\titleformat{\section}{\normalfont\bfseries\Large}{\thesection}{1em}{}
\titlespacing*{\section}{0pt}{*3}{*2}

\newtoggle{final}
\togglefalse{final} 

\newcommand{\OurAgent}{\textrm{Agent K}}

\newcommand{\version}{v1.0}

\title{\OurAgent{}~\version{}: Continual  Autonomous LLM-Agent for End-to-End Data Science}

\usepackage{tcolorbox}

\usepackage{url}      
\usepackage{hyperref}   

\definecolor{headerblue}{RGB}{250, 245, 245}
\definecolor{headerblue}{RGB}{247, 247, 247}
\definecolor{titleblack}{RGB}{80, 0, 0} 
\definecolor{abstractblack}{RGB}{80, 0, 0} 

\newtcolorbox{fullbox}{
colback=headerblue,
colframe=white,
width=\textwidth,
boxrule=0pt,
arc=10pt,
outer arc=10pt,
boxsep=10pt,
left=10pt,
right=10pt,
top=1pt,
bottom=10pt
}

\newtcolorbox{fullboxtwo}{
colback=headerblue,
colframe=white,
width=\textwidth,
boxrule=0pt,
arc=10pt,
outer arc=10pt,
boxsep=10pt,
left=10pt,
right=10pt,
top=1pt,
bottom=10pt
}

\author
{Firstname Lastname,$^{1}$ 
\\
\normalsize{$^{1}$Huawei Noah’s Ark Lab, London, UK.}\\
\normalsize{$^{2}$Technical University of Darmstadt, Darmstadt, Germany.}\\
\normalsize{$^{3}$University College London, London, UK.}\\
\\
\normalsize{
Correspondence: 
firstname.lastname@huawei.com,} \\
}

\date{}

\begin{document}




\begin{fullbox}
\vspace{1em}
\begin{center}
{\Large \textbf{Many of Your DPOs are Secretly One: Attempting Unification Through Mutual Information }}


\vspace{1em}
\textbf{\textsf{Rasul Tutunov}$^{1}$  \textsf{Antoine Grosnit}$^{1}$ and \textsf{Haitham Bou-Ammar$^{1,2}$}}\\
{\small $\ ^1$ Huawei Noah's Ark $\ ^2$ AI Centre, Department of Computer Science, UCL} 
\vspace{0.1em}

\end{center}

\noindent
\textbf{Abstract:} 
Post-alignment of large language models (LLMs) is critical in improving their utility, safety, and alignment with human intentions. Direct preference optimisation (DPO) has become one of the most widely used algorithms for achieving this alignment, given its ability to optimise models based on human feedback directly. However, the vast number of DPO variants in the literature has made it increasingly difficult for researchers to navigate and fully grasp the connections between these approaches. This paper introduces a unifying framework inspired by mutual information, which proposes a new loss function with flexible priors. By carefully specifying these priors, we demonstrate that many existing algorithms, such as SimPO, TDPO, SparsePO, and others, can be derived from our framework. This unification offers a clearer and more structured approach, allowing researchers to understand the relationships between different DPO variants better. We aim to simplify the landscape of DPO algorithms, making it easier for the research community to gain insights and foster further advancements in LLM alignment. Ultimately, we hope our framework can be a foundation for developing more robust and interpretable alignment techniques. \footnote{This is a work-in-progress document. If you believe further unifications are possible or if we have missed your algorithm, please get in touch with us at rasul.tutunov@huawei.com, and we will do our best to incorporate your suggestions.}
\end{fullbox}

\thispagestyle{fancy}

\section{Introduction}
Large language models (LLMs) exhibit remarkable language comprehension and generation abilities, allowing them to be applied in various domains, including translation \cite{brants2007large,huang2024surveylargelanguagemodels}, sentiment analysis \cite{zhang2023sentiment}, summarisation \cite{luo2024factual}, medical research \cite{thirunavukarasu2023large,zhou2023survey}, data science \cite{grosnit2024largelanguagemodelsorchestrating, Guo2024DS-Agent,Jing2024DSBenchHF}, communication networks \cite{ayed2024hermeslargelanguagemodel,boateng2024surveylargelanguagemodels,zimmer2024mixtureattentionsspeculativedecoding}, robotics and sequential decision-making \cite{christianos2023panguagentfinetunablegeneralistagent,mower2024ros,zeng2023large}. These models gain knowledge by pre-training on vast and diverse datasets containing high-quality text. During this pre-training phase, they learn to predict the next word or token in a sequence based on the context provided by the preceding tokens. This next-token prediction objective helps the model capture complex patterns, relationships, and nuances in language. 

While pre-training equips LLMs with a broad understanding of language and general world knowledge, this knowledge is often too general. It may not specialise enough for specific tasks or domains. Fine-tuning allows us to adapt these pre-trained models to particular applications by training them on smaller, task-specific datasets, improving their performance on more focused objectives. This fine-tuning stage often requires post-training. A popular method in this stage is reinforcement learning from human feedback (RLHF), which fine-tunes models using human-derived reward functions to guide responses \cite{achiam2023gpt,kaufmann2023survey,peng2023instruction,zheng2023secrets}. Despite its effectiveness, RLHF is challenging because it necessitates carefully defined reward models and precise tuning of hyperparameters, which can be resource and computationally expensive. Additionally, reward-based methods like RLHF face difficulties such as reward misalignment and out-of-distribution generalisation \cite{casper2023open, leng2024taming, wang2024secrets,yang2024bayesian}. 

In response to these challenges, reward-free approaches like direct preference optimization (DPO) have been proposed \cite{rafailov2024direct}. DPO avoids the reliance on explicit reward functions, offering a potentially more stable and efficient alternative for fine-tuning models without complex reward engineering \cite{wang2024comprehensive,xiao2024comprehensive}. However, DPO still faces significant limitations; see the exhaustive work in \cite{xiao2024comprehensive} for a more comprehensive exposition. Inspired by \cite{xiao2024comprehensive}, we note that DPO's reliance on implicit reward signals can sometimes result in policies favouring responses outside the intended distribution \cite{saeidi2024insightsalignmentevaluatingdpo,xu2024dposuperiorppollm}. Moreover, offline implementations of DPO underperform compared to online methods for aligning models \cite{ivison2024unpackingdpoppodisentangling}. Additionally, models that undergo fine-tuning via DPO may experience an ``alignment penalty'', where the alignment process reduces the model's overall effectiveness \cite{lin2024mitigatingalignmenttaxrlhf, lu2024onlinemergingoptimizersboosting}. 

Several enhanced versions of DPO have been proposed \cite{azar2023generaltheoreticalparadigmunderstand,christopoulou2024sparsepocontrollingpreferencealignment,ethayarajh2024ktomodelalignmentprospect,liu2024tisdpotokenlevelimportancesampling,meng2024simposimplepreferenceoptimization, omura2024entropycontrollabledirectpreference,zeng2024tokenleveldirectpreferenceoptimization} to address these issues. While successful in isolated instances, this expansion of closely related algorithms makes it difficult to navigate the literature and determine the most effective approach. The sheer volume of variations often leads to confusion, as many of these methods share similar foundations but differ in their specific tweaks or implementations. As such, a unified approach, where (many of) these algorithms are viewed as special cases of a more comprehensive framework, dramatically simplifies this landscape. Such an approach would allow researchers and practitioners to understand the underlying principles that drive each variation more clearly. Such a framework could lead to more efficient development and application of alignment methods, as it would help identify which elements are most important across different scenarios and how they can be combined for optimal performance. A consolidated framework would make it easier to pinpoint areas for improvement, ensuring that progress is made more effectively and with fewer redundant efforts. 

This paper addresses this challenge by proposing a unified framework explaining why many current algorithms can be seen as special cases of a broader paradigm. We accomplish this by relying on a more general constraint reinforcement learning framework that incorporates learnable priors and is principally motivated from a \emph{mutual information} perspective. This approach gives rise to a novel DPO method class titled Mutual information DPO or MI-DPO for short. MI-DPO requires prior specifications that can be optimised or manually pre-defined. By defining these priors, we, surprisingly, observe that we can unify various existing algorithms, revealing how they naturally emerge as instantiations from our framework. Furthermore, we demonstrate that \emph{directly optimising} for the prior leads to improved loss function outcomes around optima, offering a more promising approach to model alignment. In short, our contributions can be summarised as: 
\begin{itemize}
    \item Proposing a more general formulation - titled MI-DPO (mutual information DPO) - based on learnable priors for preference alignment inspired by mutual information;  
    \item Showing that various algorithms in literature naturally arise as special cases with the relevant prior choice; and 
    \item Theoretically discussing MI-DPO and demonstrating its advantages compared to current approaches.  
\end{itemize}
\begin{fullboxtwo} 
\begin{center}
{\underline{\textbf{A summary presenting the prior choices and the algorithms recovered.}}}
\end{center}
\begin{align*}
   \text{DPO}&\rightarrow \textcolor{black}{\zeta \equiv \pi_{\text{ref}}(y|x)}  \\
    \text{DICE} &  \rightarrow \textcolor{black}{\zeta \equiv \pi^{(t-1)}_{\text{LLM}}(y|x)} \\
\text{cEntropy} &\rightarrow \textcolor{black}{\zeta \propto \pi_{\text{ref}}(y|x)[\pi_{\text{LLM}}(y|x)]^{1-\eta}}\\
   \text{R-DPO} &\rightarrow\textcolor{black}{\zeta \propto \frac{\pi_{\text{ref}}(y|x)}{\exp(\beta |y|)}}\\
    \text{simPO} &\rightarrow \textcolor{black}{\zeta \propto \left[\pi_{\text{LLM}}(y|x)\right]^{\left(1-\frac{1}{|y|}\right)}\exp \beta(y,x)}  \\
      \text{TDPO} &\rightarrow \textcolor{black}{\zeta \propto \pi_{\text{ref}}(y|x) \exp\left(-\sum_{t=1}^{|y|}\text{KL}\left(\pi_{\text{ref}}\left(\cdot|x, y^{<t}\right)||\pi_{\text{LLM}}\left(\cdot|x, y^{<t}\right) \right)\right)}\\ 
      \text{TIS-DPO} &\rightarrow \zeta \propto
    \pi_{\text{LLM}}(y|x)\prod_{i=1}^{|y|}\left(\frac{\pi_{\text{ref}}(y_i|x,y_{<i})}{\pi_{\text{LLM}}(y_i|x,y_{<i})}\right)^{w(y_i;x)}\\
    & \hspace{12em}\exp\Bigg(D_{\text{SeqL}}(x,y, w(y;x); \pi_{\text{LLM}}||\pi_{\text{ref}})\Bigg) \\ 
     \text{sparsePO} &\rightarrow \zeta \propto  \exp\Bigg(\sum_{t=1}^{|y|}\log \frac{[\pi_{\text{LLM}}(\cdot|x,y^{<t})]^{1-\mu_1 (y_t)}}{\pi_{\text{ref}}(.|x,y^{<t})^{-\mu_1(y^t)}} \\ 
     & \hspace{7em}- \sum_{t=1}^{|y|}\mu_2\left(y^{t}\right) \text{KL}\left(\pi_{\text{ref}}(\cdot|x, y^{<t})||\pi_{\text{LLM}}(\cdot|x, y^{<t})\right)\Bigg) 
\end{align*}
\end{fullboxtwo}

\section{Direct Preference Optimisation}
DPO \cite{rafailov2024direct} was introduced as a reward-free approach to model alignment, aiming to overcome the challenges associated with traditional reinforcement learning methods, such as the need for complex reward engineering and high computational costs. Unlike RLHF, which relies on predefined reward functions to guide the model's behaviour, DPO leverages human preferences directly to align model outputs with user expectations. 

At a high level, DPO optimises a model to prefer responses that align with human judgment over others without the need for explicit reward signals. Instead of computing rewards for every action, DPO focuses on ranking pairs of responses based on human feedback and then directly optimises the model to select responses that are more likely to be ranked higher. Formally, given a dataset of winning and losing answers $\mathcal{D}=\{x^{(i)}, y^{(i)}_w, y^{(i)}_l\}_{i=1}^{n}$, DPO minimises the following objective: 
\begin{equation}
\label{Eq:DPO}
    \mathcal{J}_{\text{DPO}}\left(\pi_{\text{LLM}}\right) = - \mathbb{E}_{(x, y_w, y_l) \sim \mathcal{D}}\left[\log \text{sigmoid}\left(\alpha \log \frac{\pi_{\text{LLM}}(y_w|x)}{\pi_{\text{ref}}(y_w|x)} - \alpha \log \frac{\pi_{\text{LLM}}(y_l|x)}{\pi_{\text{ref}}(y_l|x) } \right) \right],
\end{equation}
with $\pi_{\text{ref}}(y|x)$ being a reference frozen LLM policy. 

Intuitively, the objective in Equation \ref{Eq:DPO} compares the likelihood of two responses $y_w$ (a preferred response) and $y_l$ (a less preferred response), under the model being optimised $\pi_{\text{LLM}}$ and a reference model $\pi_{\text{ref}}$. The goal is to maximise the preference alignment of the model's outputs while penalising deviations from the reference model.

\paragraph{Deriving DPO's Objective:} We now present DPO's standard derivation that we extend to general priors in Section \ref{Sec:GDPO}. As noted in \cite{rafailov2024direct}, Having pre-trained an LLM, the model is prompted with various prompts $x$ to produce a pair of responses. Those are then presented to human annotators to rank one answer over the other according to their preferences. These preferences are assumed to be generated from a latent reward such that: 
\begin{align}
\label{Eq:BT}
   (\text{Bradley-Terry Model}) \ \  \textbf{Pr}(y_w \succ y_l|x) &= \frac{\exp(r(x,y_w))}{\exp(r(x,y_w)) + \exp(r(x,y_l))}, 
\end{align}
where $y_w$ is the pre-trained model's response preferred by the users to $y_l$, and $x$ is a prompt. 

Assuming access to the human-annotated dataset $\mathcal{D}$, we can parameterise a model to train latent rewards. Once acquired, one applies reinforcement learning in post-training on top of the learnt reward to align the pre-trained LLM. Since learning this reward is challenging, DPO followed prior work \cite{peters2007reinforcement} to eliminate rewards altogether by relying on closed-form policies. 

In details, the story starts with KL-regularised reinforcement learning \cite{jaques2017sequencetutorconservativefinetuning,peters2008natural,schulman2015trust} that solves the following problem:   
\begin{equation}
\label{Eq:KLReg}
    \max_{\pi_{\text{LLM}}} \mathbb{E}_{x, y \sim \pi_{\text{LLM}}(y|x)}\left[r(x,y) - \frac{1}{\beta} \log \frac{\pi_{\text{LLM}}(y|x)}{\pi_{\text{ref}}(y|x)}\right],
\end{equation}
where $\beta$ is a regularisation constant trading-off rewards and ``closeness'' to $\pi_{\text{ref}}(y|x)$. It is easy to see that the optimal policy of the problem in Equation \ref{Eq:KLReg} exhibits a closed form as follows: 
\begin{equation*}
    \pi_{\text{LLM}}^{\star}(y|x) = \frac{1}{Z(x)} \pi_{\text{ref}}(y|x) \exp \left(\frac{1}{\beta} r(x,y)\right),
\end{equation*}
with $Z(x) = \sum_{y^{\prime}} \pi_{\text{ref}}(y^{\prime}|x)\exp\left(\frac{1}{\beta} r (x,y^{\prime})\right)$ being a normalisation constant that is highly challenging to compute. From $\pi^{\star}_{\text{LLM}}(y|x)$, we can further say that: 
\begin{equation*}
    r(x,y) = \beta \log \frac{\pi_{\text{LLM}}^{\star}(y|x)}{\pi_{\text{ref}}(y|x)} + \beta \log Z(x).
\end{equation*}
Fortunately, upon plugging this equation back into the Bradley-Terry model (Equation \ref{Eq:BT}) and taking logarithms, the normalisation constant vanishes, recovering $\mathcal{J}_{\text{DPO}}(\pi_{\text{LLM}})$ from Equation \ref{Eq:DPO}. 

\section{Mutual Information DPO} \label{Sec:GDPO}
This section departs from the standard DPO derivation by introducing general priors. We follow a similar principle to DPO to derive our loss in that we aim to circumvent modelling latent reward functions. 
\subsection{Formulation \& Motivation}
Similar to the works in \cite{grau2018soft,ma2024mutual,mazoure2020deep, tschannen2019mutual}, we consider the following generalisation of regularised reinforcement learning, where we assume that we want our policy $\pi_{\text{LLM}}$ not to deviate from a prior $\zeta(y)$:
\begin{align*}
    \max_{\pi_{\text{LLM}}} \mathbb{E}_{x, y \sim \pi_{\text{LLM}}(y|x)}\left[r(x,y) - \frac{1}{\beta} \log \frac{\pi_{\text{LLM}}(y|x)}{\zeta(y)}\right] &= \max_{\pi_{\text{LLM}}}\sum_{x, y} p(x)\pi_{\text{LLM}}(y|x) r(x,y) \\ 
    & \hspace{5em} - \frac{1}{\beta} \sum_{x} p(x) \text{KL}\left(\pi_{\text{LLM}}(y|x)||\zeta(y)\right),
\end{align*}
with $p(x)$ being a distribution over input questions (e.g., uniform from a dataset), the KL refers to the Kullback-Leiber divergence. Similar to \cite{grau2018soft}, we propose to go one step further and also optimise for $\zeta(\cdot)$ in addition to $\pi_{\text{LLM}}(\cdot)$. This implies that the policies are guided to stay close to an optimal prior distribution. As a result, we formalise the optimisation problem in a way that balances aligning the model's behaviour with human preferences while maintaining consistency with this prior distribution. This leads to the following optimisation problem: 
\begin{align*}
    \max_{\pi_{\text{LLM}}, \zeta(y)}\sum_{x, y} p(x)\pi_{\text{LLM}}(y|x) r(x,y) - \frac{1}{\beta} \sum_{x} p(x) \text{KL}\left(\pi_{\text{LLM}}(y|x)||\zeta(y)\right). 
\end{align*}

Since the first term $\sum_{x, y} p(x)\pi_{\text{LLM}}(y|x) r(x,y)$ is independent of $\zeta(y)$, we can rewrite the above equation as: 
\begin{align*}
    \max_{\pi_{\text{LLM}}}\sum_{x, y} p(x)\pi_{\text{LLM}}(y|x) r(x,y) + \max_{\zeta(y)} \left\{- \frac{1}{\beta} \sum_{x} p(x) \text{KL}\left(\pi_{\text{LLM}}(y|x)||\zeta(y)\right)\right\}. 
\end{align*}
We can further rewrite this equation by minimising over $\zeta(y)$ such that: 

\begin{equation}
\label{Eq:Min}
  \max_{\pi_{\text{LLM}}}\left\{\sum_{x, y} p(x)\pi_{\text{LLM}}(y|x) r(x,y) -\frac{1}{\beta}\min_{\zeta(y)} \left\{ \sum_{x} p(x) \text{KL}\left(\pi_{\text{LLM}}(y|x)||\zeta(y)\right)\right\}\right\}. 
\end{equation} 
Equation \ref{Eq:Min} is intuitive. It aims to maximise rewards while simultaneously encouraging the model to remain close to an \emph{optimal prior distribution}. Contrary to current techniques, this optimal prior is not fixed but is itself learnable, allowing the framework to adapt flexibly. By balancing these two objectives—reward maximisation and adherence to the prior—the model avoids overfitting to specific preferences while maintaining general alignment with desirable behaviour. 

\paragraph{Motivation from Mutual Information:} While this formulation is intuitive, it is also grounded in a more principled framework, as we demonstrate next. Specifically, it can be motivated from a \emph{mutual information} perspective, which provides a theoretical basis for balancing the trade-off between reward maximisation and adherence to the prior. Before doing so, we need first to present a Lemma from \cite{thomas2006elements} that has also been widely used in literature (e.g., by the work in \cite{grau2018soft}): 

\begin{lemma}[Mutual Information \cite{thomas2006elements}]
Let $I_{g}$ be a functional defined as: 
\begin{equation*}
    I_{g}(p_{X}, p_{Y|X}, q_Y) = \sum_x p_{X}(x)\text{KL}\left(p_{Y|X}(\cdot|x)||q_{Y}(\cdot)\right),
\end{equation*}
where $p_{X}(x)$ is the distribution of the input, $p_{Y|X}(y|x)$ is the conditional distribution of the output conditioned on the input, and $q_{Y}(y)$ a variational distribution of the output. Furthermore, define the mutual information as: 
\begin{equation*}
    I(X;Y) = \sum_{x,y} p(x,y) \log \frac{p(x,y)}{p_{Y}(y)p_{X}(x)} = \sum_{x} p_{X}(x)\text{KL}\left(p_{Y|X}(\cdot|x)||p_{Y}(y)\right). 
\end{equation*}
The mutual information is recovered when 
\begin{equation*}
    I[X;Y] = \min_{q_Y} I_g(p_{X}, p_{Y|X},q_{Y}).
\end{equation*}
Furthermore, the optimal variational distribution is given by: 
\begin{equation*}
    q^{\star}_Y(y) = \sum_{x} p_{X}(x)p_{Y|X}(y|x) \ \ \text{(the true marginal distribution)}. 
    \end{equation*} 
\end{lemma} 
The above lemma is also intuitive and easy to understand. It simply introduces a mutual-information-based perspective on optimisation. In detail, it provides a way to understand mutual information as the minimum KL divergence between the true condition distribution $p_{Y|X}$ and a variational distribution $q_Y$ that approximates it. In other words, it says that mutual information $I(X,Y)$ measures how well the output $Y$ can be predicted from the input $X$ by finding the best possible approximation $q_Y$. This lemma highlights that the closer $q_Y$ is to $p_{Y|X}$, the better it captures the true dependency between $X$ and $Y$. Furthermore, the optimal $q_Y$ is shown to be the true marginal of $Y$ conditioned on $X$, ensuring that the variational approximation aligns with the underlying data distribution.
    
Now, we can apply the above Lemma to our problem, specifically to the following term from Equation \ref{Eq:Min}: 
\begin{equation*}
    -\frac{1}{\beta}\min_{\zeta(y)} \left\{ \sum_{x} p(x) \text{KL}\left(\pi_{\text{LLM}}(y|x)||\zeta(y)\right)\right\} = -\frac{1}{\beta}I[X;Y]. 
\end{equation*}
This allows us to write a penalty on the mutual information between states and actions, yielding the following RL-constrained objective: 
\begin{align*}
      \boxed{\max_{\pi_{\text{LLM}}}\sum_{x, y} p(x)\pi_{\text{LLM}}(y|x) r(x,y) -\frac{1}{\beta} I[X;Y].}
\end{align*}
While we introduce such a problem in the context of LLM alignment, we are not the first to study such an objective in reinforcement learning. In fact, this has been rigorously analysed in the works of \cite{grau2018soft,tschannen2019mutual}, where it has been pointed out that this problem is mathematically equivalent to rate-distortion from information theory \cite{shannon1959coding}. Rate distortion formulates how to efficiently send information over an information-theoretic channel with a limited transmission rate. In a decision-making context, the agent is considered an information-theoretic channel $\pi_{\text{LLM}}(y|x)$ where $x$ is the channel input and $y$ its output. The agent's objective is to maximise the expected reward while operating under a constraint on how much information it can transmit. The mutual information between the input prompts and the generated answers defines this transmission limit. Intuitively, the agent must focus only on the relevant information to maximise the reward, discarding any extraneous or irrelevant details. By doing so, the agent ensures it stays within the allowed information transmission rate, effectively balancing the trade-off between using information efficiently and achieving high rewards; see the work in \cite{grau2018soft} for a more detailed exposition.

\subsection{Deriving Mutual Information DPO}
Deriving the new objective for mutual information DPO (MI-DPO) under a general prior is rather straightforward, following the same two phases of standard DPO: 1) arriving at an optimal solution for $\pi^{\star}_{\text{LLM}}$, and 2) plugging the resultant reward in the Bradley-Terry model from Equation \ref{Eq:BT}. This section presents those two phases. Here, we assume $\zeta(y)$ is free of optimisation since we wish to recover special cases, as shown in Section \ref{Sec:SpecialCases}. Later on, in Section \ref{Sec:GeneralTheory}, we analyse the benefits of optimising $\zeta(y)$. 

\paragraph{Optimal Policies Under General Priors:} This section provides a sketch of the derivations with their comprehensive form shown in Appendix \ref{App:DerivingDuals}. Having $\zeta(y)$ free of optimisation, we consider the following problem: 
\begin{equation*}
    \mathcal{J}(\pi_{\text{LLM}}) = \max_{\pi_{\text{LLM}}} \mathbb{E}_{x, y \sim \pi_{\text{LLM}}(y|x)}\left[r(x,y) - \frac{1}{\beta} \log \frac{\pi_{\text{LLM}}(y|x)}{\zeta(y)}\right]  \ \ \overbrace{\text{s.t.} \sum_{y} \pi_{\text{LLM}}(y|x) = 1}^{\text{a valid probability distribution}},
\end{equation*}
where we added the constraint that $\pi_{\text{LLM}}(y|x)$ should be a valid probability distribution when summed over all possible answers $y$. Given a fixed $x$, let us now consider the following objective: 
\begin{align*}
    \mathcal{J}_{{x}}(\pi_{\text{LLM}})&= \mathbb{E}_{y \sim \pi_{\text{LLM}}(y|x)}\left[r(x,y) - \frac{1}{\beta}\log \pi_{\text{LLM}}(y|x)+\frac{1}{\beta}\log\zeta(y)\right] \\
    & = \sum_{y} \pi_{\text{LLM}}(y|x)r(x,y) - \frac{1}{\beta} \sum_{y} \pi_{\text{LLM}}(y|x) \log  \pi_{\text{LLM}}(y|x) + \frac{1}{\beta} \sum_{y} \pi_{\text{LLM}}(y|x)\log \zeta(y) \\
    & = \sum_{y}  \pi_{\text{LLM}}(y|x)\left[r(x,y)+\frac{1}{\beta}\log \zeta(y)\right]+\frac{1}{\beta}\mathcal{H}\left( \pi_{\text{LLM}}(y|x)\right),
\end{align*}
with the constraint that $\sum_{y}\pi_{\text{LLM}}(y|x) = 1$, and $\mathcal{H}\left( \pi_{\text{LLM}}(y|x)\right)$ denoting the entropy of the policy $\pi_{\text{LLM}}(y|x)$. 

Since the above objective represents a constrained optimisation problem, we transform it into an unconstrained one by writing the Lagrangian, yielding: 
\begin{align*}
    \mathcal{J}_{\text{lag}}(\pi_{\text{LLM}}(y|x), \lambda) &= \sum_{y}  \pi_{\text{LLM}}(y|x)\left[r(x,y)+\frac{1}{\beta}\log \zeta(y)\right]  \\ 
    &\hspace{13em}+\frac{1}{\beta}\mathcal{H}\left( \pi_{\text{LLM}}(y|x)\right) + \lambda(\sum_{y}\pi_{\text{LLM}}(y|x) - 1).
\end{align*}
Taking the gradient of $\mathcal{J}_{\text{lag}}$ with respect to $\pi_{\text{LLM}}(y|x)$ and setting it to zero gives us the form of the optimal policy as a function of the dual variables $\lambda$: 
\begin{align*}
    \pi^{\star}_{\text{LLM}}(y|x) = \zeta(y)\text{exp}(\beta r(x,y) -1)\exp(\beta\lambda).
\end{align*}

Upon substituting this result back into the Lagrangian and deriving optimising the dual variables $\lambda$, we can show that: 
\begin{align}
\label{Eq:ZetaFun}
    \pi^{\star}_{\text{LLM}}(y|x) = \frac{\zeta(y)\exp(\beta r(x,y))}{\hat{\mathcal{Z}(x)}} \ \ \text{with $\hat{\mathcal{Z}}=\sum_{y^{\prime}}\zeta\left(y^{\prime}\right)\exp\left(\beta r(x,y^{\prime})\right).$}
\end{align}
Intuitively, this policy is similar to the standard optimal policy used in traditional DPO.  However, the key difference is that it introduces a more general prior distribution instead of relying on a fixed reference policy. The prior can be specified or adapted to represent different assumptions or preferences about the output space, which, when chosen appropriately, recovers many existing variants of DPO currently available in the literature; see Section \ref{Sec:SpecialCases}. 

\paragraph{Generalised DPO's Loss:} To derive our new DPO loss, we can follow the second phase of the standard approach as presented in \cite{rafailov2024direct}. In this phase, we need to rewrite rewards as a function of $\pi^{\star}_{\text{LLM}}$ and plug in our resultant in Bradley-Terry's model. From Equation \ref{Eq:ZetaFun}, we see that $r(x,y)$ exhibits the following form: 
\begin{align*}
\exp(\beta r(x,y)) &= \hat{\mathcal{Z}}\frac{\pi_{\text{LLM}}^{\star}(y|x)}{\zeta(y)} \implies r(x,y) = \frac{1}{\beta}\log\left[\hat{\mathcal{Z}}\frac{\pi_{\text{LLM}}^{\star}(y|x)}{\zeta(y)}\right]   \\
 r(x,y) &= \frac{1}{\beta}\left[\log \pi_{\text{LLM}}^{\star}(y|x) + \log \hat{\mathcal{Z}} - \log \zeta(y)\right] = \frac{1}{\beta} \log \frac{\pi_{\text{LLM}}^{\star}(y|x)}{\zeta(y)} + \frac{1}{\beta} \log \hat{\mathcal{Z}}. 
\end{align*}
Now, let us substitute those into the Bradley-Terry's model: 
\begin{align*}
\textbf{Pr}(y_1 \succ y_2|x) &= \frac{\exp\left(\frac{1}{\beta} \log \frac{\pi_{\text{LLM}}^{\star}(y_{1}|x)}{\zeta(y_{1})} + \frac{1}{\beta} \log \hat{\mathcal{Z}}\right)}{\exp\left(\frac{1}{\beta} \log \frac{\pi_{\text{LLM}}^{\star}(y_1|x)}{\zeta(y_1)} + \frac{1}{\beta} \log \hat{\mathcal{Z}}\right) + \exp\left(\frac{1}{\beta} \log \frac{\pi_{\text{LLM}}^{\star}(y_2|x)}{\zeta(y_2)} + \frac{1}{\beta} \log \hat{\mathcal{Z}}\right)} \\
& = \frac{\exp\left(\frac{1}{\beta} \log \hat{\mathcal{Z}}\right) \exp\left(\frac{1}{\beta}\log\frac{\pi_{\text{LLM}}^{\star}(y_1|x)}{\zeta(y_1)}\right)}{\exp\left(\frac{1}{\beta} \log \hat{\mathcal{Z}}\right)\left[\exp\left(\frac{1}{\beta}\log\frac{\pi_{\text{LLM}}^{\star}(y_1|x)}{\zeta(y_1)}\right)+\exp\left(\frac{1}{\beta}\log\frac{\pi_{\text{LLM}}^{\star}(y_2|x)}{\zeta(y_2)}\right)\right]} \\
 & = \text{sigmoid}\left(\alpha \log \frac{\pi_{\text{LLM}}^{\star}(y_1|x)}{\zeta(y_1) } - \alpha \log \frac{\pi_{\text{LLM}}^{\star}(y_2|x)}{\zeta(y_2) } \right),
\end{align*}
with $\alpha = \frac{1}{\beta}$.

Given a data set of preferred and less preferred answers $y_w$ and $y_l$, and taking the logarithm of the above gives our new loss: 
\begin{equation*}
   \boxed{ \mathcal{J}_{\text{MI-DPO}}\left(\pi_{\text{LLM}}, \zeta\right) = - \mathbb{E}_{(x, y_w, y_l) \sim \mathcal{D}}\left[\log \text{sigmoid}\left(\alpha \log \frac{\pi_{\text{LLM}}(y_w|x)}{\zeta(y_w) } - \alpha \log \frac{\pi_{\text{LLM}}(y_l|x)}{\zeta(y_l) } \right) \right]}
\end{equation*}

\section{Recovering Special Cases} 
\label{Sec:SpecialCases}
Now that we have introduced the general prior formulation, we focus on its practical utility by exploring specific cases. This section will demonstrate how selecting appropriate prior distributions allows us to recover various existing algorithms as special cases. Specifically, we will show how this framework unifies eight state-of-the-art preference optimisation algorithms. We provide an overview of the key results, with the complete derivations detailed in Appendix \ref{App:SpecialCases} for reference.

The unification process is straightforward. To unify a given algorithm within our framework, we identify its loss function and determine the prior distribution that would make the ratios inside our sigmoid expression match those in the target algorithm. This strategy works well for symmetric cases where the prior applies equally to preferred and non-preferred answers. However, we can extend our formulation further by introducing additional constraints for more general scenarios—such as those involving global regularisers or asymmetrical constraints; see Section \ref{Sec:GeneralTheory}. 

We start with two simple, special cases: standard DPO (Equation \ref{Eq:DPO}) and DICE-based (iterative-like) DPO \cite{chen2024bootstrappinglanguagemodelsdpo} where the loss functions for any iteration $t$ is written as: 
\begin{equation*}
    \mathcal{J}_{\text{DICE}}(\pi_{\text{LLM}}) = - \mathbb{E}_{(x, y_w, y_l) \sim \mathcal{D}_t}\left[\log \text{sigmoid}\left(\alpha \log \frac{\pi_{\text{LLM}}(y_w|x)}{\pi^{(t-1)}_{\text{LLM}}(y_w|x) } - \alpha \log \frac{\pi_{\text{LLM}}(y_l|x)}{\pi^{(t-1)}_{\text{LLM}}(y_l|x) } \right) \right],
\end{equation*}
where $\pi^{(t-1)}_{\text{LLM}}(y|x)$ is the previous iteration updated policy. 

It is easy to see that if we set $\zeta$ to $\zeta_{\text{DPO}} \equiv \pi_{\text{ref}}(y|x)$ and to $\zeta_{\text{DICE}}(y) = \pi_{\text{LLM}}^{(t-1)}\left(y|x\right)$, we recover both algorithms as special cases: 
\begin{equation*}
   \boxed{
    \mathcal{J}_{\text{MI-DPO}}\left(\pi_{\text{LLM}}, \textcolor{red}{\zeta \equiv \zeta_{\text{DPO}}}\right)
     \rightarrow \mathcal{J}_{\text{DPO}}(\pi_{\text{LLM}}) } \ \ \ \ \ \ \   \boxed{
    \mathcal{J}_{\text{MI-DPO}}\left(\pi_{\text{LLM}}, \textcolor{red}{\zeta \equiv \zeta_{\text{DICE}}}\right)
     \rightarrow \mathcal{J}_{\text{DICE}}(\pi_{\text{LLM}})}
\end{equation*}
\paragraph{Recovering Entropy Controllable DPO:} Entropy controllable DPO \cite{omura2024entropycontrollabledirectpreference} is a simple extension to standard DPO incorporating a parameter $\eta$, allowing for entropy adjustment to aid in capturing modes of reference policy distributions. Its loss is defined by: 
\begin{equation*}
    \mathcal{J}_{\text{cEntopy}}(\pi_{\text{LLM}}) = - \mathbb{E}_{(x, y_w, y_l) \sim \mathcal{D}_t}\left[\log \text{sigmoid}\left(\alpha \log \frac{\pi^{\eta}_{\text{LLM}}(y_w|x)}{\pi_{\text{ref}}(y_w|x) } - \alpha \log \frac{\pi^{\eta}_{\text{LLM}}(y_l|x)}{\pi_{\text{ref}}(y_l|x) } \right) \right].
\end{equation*}
Setting $\zeta$ to $\zeta_{\text{cEntropy}}(y) = \pi_{\text{ref}}(y|x)[\pi_{\text{LLM}}(y|x)]^{1-\eta}$, we easily recover this special case, whereby: 
\begin{equation*}
   \boxed{ \mathcal{J}_{\text{MI-DPO}}(\pi_{\text{LLM}}, \textcolor{red}{\zeta \propto \zeta_{\text{cEntropy}}}) \rightarrow \mathcal{J}_{\text{cEntropy}}(\pi_{\text{LLM}})}
\end{equation*}

\paragraph{Recovering SimPO:} SimPO \cite{meng2024simposimplepreferenceoptimization} is an algorithm that replaces DPO's reliance on a reference model with a reference-free reward based on the average log probability of a response, aligning the reward function with the metric used for generation. This modification reduces computational and memory overhead while maintaining strong performance. SimPO also incorporates a target reward margin into the Bradley-Terry objective, encouraging a larger margin between preferred and non-preferred responses, further enhancing model performance. The standard loss for SimPO is given by: 
\begin{align*}
    \mathcal{J}_{\text{SimPO}} (\pi_{\text{LLM}}) &= - \mathbb{E}_{(x, y_w, y_l) \sim \mathcal{D}}\Bigg[\log \text{sigmoid}\Bigg(\frac{\alpha}{|y_w|}\log \pi_{\text{LLM}}(y_w|x) \\
    & \hspace{20em}- \frac{\alpha}{|y_l|}\log \pi_{\text{LLM}}(y_l|x)-\gamma\Bigg)\Bigg],  
\end{align*}
where $|y|$ denotes the length of an answer $y$, and $\alpha$ and $\gamma$ are constants. 

We can recover $\mathcal{J}_{\text{simPO}}$ from $\mathcal{J}_{\text{MI-DPO}}$ under the additional assumption that ``winning'' and ``losing'' labels for the answers in $\mathcal{D}_t$ are based on an absolute criterion. This  corresponds to standard fine-tuning scenarios where examples are valid/invalid solutions to a mathematical problem, or are correct/incorrect codes associated with a coding problem prompt.
Formally, this assumption states that there is no $x$ such that both $(x, y_1, y_2) \in \mathcal{D}$ and $(x, y_2, y_3) \in \mathcal{D}$ hold simultaneously for different $y_1, y_2, y_3$.
This assumption is necessary to pick a $\zeta$ function treating differently winning and losing responses, as follows: 
\begin{align*}
    &\zeta_{\text{simPO}}(y) \propto \left[\pi_{\text{LLM}}(y|x)\right]^{\left(1-\frac{1}{|y|}\right)}\exp{\beta(y,x)}, \ \ \text{with} \ \ 
    \beta(y,x) = \begin{cases}
        \frac{\gamma}{2\alpha} & \text{if } \ y = y_w \ \ \text{for } \ x,\\
        -\frac{\gamma}{2\alpha} & \text{if } \  y = y_l \ \ \text{ for } \ x.\\
    \end{cases}
\end{align*}
Using the above, we recover the work from \cite{meng2024simposimplepreferenceoptimization} since: 
\begin{equation*}
\boxed{    \mathcal{J}_{\text{MI-DPO}}\left(\pi_{\text{LLM}}(y|x), \textcolor{red}{\zeta \propto \left[\pi_{\text{LLM}}(y|x)\right]^{\left(1-\frac{1}{|y|}\right)}\exp \beta(y,x)}\right) \rightarrow \mathcal{J}_{\text{simPO}}(\pi_{\text{LLM}})} 
\end{equation*}

\paragraph{Recovering R-DPO:} R-DPO explores the issue of DPO tending to produce excessively long responses that deviate from user preferences \cite{park2024disentanglinglengthqualitydirect}. To address this, RDPO proposes a length-regularised version of DPO that introduces a penalty term: 
\begin{align*}
    \mathcal{J}_{\text{R-DPO}}(\pi_{\text{LLM}}) &=-\mathbb{E}_{(x,y_w, y_l) \sim \mathcal{D}} \Bigg[\log \text{sigmoid}\Bigg(\alpha \log \frac{\pi_{\text{LLM}}(y_w|x)}{\pi_{\text{ref}}(y_w|x)} - \alpha \log \frac{\pi_{\text{LLM}}(y_l|x)}{\pi_{\text{ref}}(y_l|x)} \\ 
    &\hspace{25em} + \underbrace{\Bigg(\tilde{\alpha} |y_w| - \tilde{\alpha} |y_l|\Bigg)}_{\text{regulariser}}
    \Bigg)\Bigg].
\end{align*}
Such a choice of $\zeta$, gives us yet another special case of $\mathcal{J}_{\text{MI-DPO}}$: 
\begin{equation*}
\boxed{
    \mathcal{J}_{\text{MI-DPO}}\left(\pi_{\text{LLM}}(y|x), \textcolor{red}{\zeta \propto \frac{\pi_{\text{ref}}(y|x)}{\exp(\beta |y|)}}\right) \rightarrow \mathcal{J}_{\text{R-DPO}}} 
\end{equation*}

\paragraph{Recovering Token-Level DPO:} The token-level direct preference optimisation (TDPO) is a novel approach to aligning large language models (LLMs) with human preferences by optimising them at the token level rather than the sentence level \cite{zeng2024tokenleveldirectpreferenceoptimization}. TDPO improves existing methods by incorporating forward KL divergence constraints for each token, enabling better control of alignment and generation diversity. TDPO's loss function is given by:  
\begin{align*}
    \mathcal{J}_{\text{TDPO}}(\pi_{\text{LLM}}) &= -\mathbb{E}_{(x, y_w, y_l)\sim \mathcal{D}}\Bigg[\log \text{sigmoid}\Bigg(\alpha \log \frac{\pi_{\text{LLM}}(y_w|x)}{\pi_{\text{ref}}(y_w|x)} - \alpha \log \frac{\pi_{\text{LLM}}(y_l|x)}{\pi_{\text{ref}}(y_l|x)} \\ 
   & \hspace{27em} \underbrace{-\alpha\Delta(x, y_w, y_l)}_{\text{regulariser}}\Bigg)\Bigg], 
\end{align*}
with $\Delta(x, y_w, y_l)$ defined as follows: 
\begin{equation*}
    \Delta(x, y_w, y_l) = \sum_{t=1}^{|y_l|} \text{KL}\left(\pi_{\text{ref}}(\cdot|[x, y_{l}^{<t}])||\pi_{\text{LLM}}(\cdot|[x, y_{l}^{<t}])\right) - \sum_{t=1}^{|y_w|} \text{KL}\left(\pi_{\text{ref}}(\cdot|[x, y_{w}^{<t}])||\pi_{\text{LLM}}(\cdot|[x, y_{w}^{<t}])\right).
\end{equation*}
Now, if we are to set our prior as follows: 
\begin{equation*}
    \zeta_{\text{TDPO}}(y) \propto \pi_{\text{ref}}(y|x) \exp\left(-\sum_{t=1}^{|y|}\text{KL}\left(\pi_{\text{ref}}\left(\cdot|x, y^{<t}\right)||\pi_{\text{LLM}}\left(\cdot|x, y^{<t}\right) \right)\right),
\end{equation*}
we recover we recover $\mathcal{J}_{\text{TDPO}}(\pi_{\text{LLM}})$ as a special case of the formulation from $\mathcal{J}_{\text{MI-DPO}}(\pi_{\text{LLM}})$ since: 
\begin{equation*}
   \boxed{ \mathcal{J}_{\text{MI-DPO}}\left(\cdot, \textcolor{red}{\zeta \propto \pi_{\text{ref}}(y|x) \exp\left(-\sum_{t=1}^{|y|}\text{KL}\left(\pi_{\text{ref}}\left(\cdot|x, y^{<t}\right)||\pi_{\text{LLM}}\left(\cdot|x, y^{<t}\right) \right)\right)}\right) \rightarrow \mathcal{J}_{\text{TDPO}}\left(\cdot\right)} 
\end{equation*}

\paragraph{Recovering Token-Level Importance Sampled DPO:} The TIS-DPO (token-level importance sampling for direct preference optimisation) algorithm addresses a key limitation in standard DPO: the uniform treatment of all tokens, which fails to account for differences in their importance \cite{liu2024tisdpotokenlevelimportancesampling}. TIS-DPO assigns weights to tokens based on their relevance to the reward, yielding the following loss function: 
\begin{equation*}
    \mathcal{J}_{\text{TIS-DPO}} = -\mathbb{E}_{(x,y_w,y_l)\sim\mathcal{D}}\left[\log\text{sigmoid}\left(u(x,y_w,y_{l},\pi_{\text{LLM}}, w^w, w^l ) - \eta(x,y_w,y_{l},\pi_{\text{LLM}}, w^w, w^l )\right)\right],
\end{equation*}
such that $u(x,y_w,y_{l},\pi_{\text{LLM}}, w^w, w^l )$ is defined as: 
\begin{align*}
    u(x,y_w,y_{l},\pi_{\text{LLM}}, w^w, w^l ) = \sum_{i=1}^{T_w}w^w_i\alpha\log\frac{\pi_{LLM}(y_{w_i}|x, y_{w_{<i}})}{\pi_{\text{ref}}(y_{w_i}|x,y_{w_{<i}})} - \sum_{j=1}^{T_l}w^l_j\alpha\log\frac{\pi_{LLM}(y_{l_j}|x, y_{l_{<j}})}{\pi_{\text{ref}}(y_{l_j}|x,y_{l_{<j}})}.
\end{align*}
Furthermore, the second term $\eta(x,y_w,y_{l},\pi_{\text{LLM}}, w^w, w^l )$ is an altered version of TDPO's sequential Kullback-Leibler component: 
\begin{align*}
    \eta(x,y_w,y_{l},\pi_{\text{LLM}}, w^w, w^l ) = \alpha D_{\text{SeqL}}(x,y_w, w^w; \pi_{\text{LLM}}||\pi_{\text{ref}}) - \alpha D_{\text{SeqL}}(x,y_l, w^l; \pi_{\text{LLM}}||\pi_{\text{ref}}).
\end{align*}
In the above, $w^w_i$ and $w^l_j$ are weights  that are applied token-level. Finally, $D_{\text{SeqL}}(x,y_l, w^l; \pi_{\text{LLM}}||\pi_{\text{ref}})$ is weighted sequence KL-divergence defined as:
\begin{align*}
    &D_{\text{SeqL}}(x,y, \tilde{w}; \pi_{1}||\pi_{2}) = \sum_{i=1}^{|y|}\tilde{w}_i\text{KL}(\pi_1(y_i| x, y_{<i}) || \pi_2(y_i| x, y_{<i}))
\end{align*}
To recover $\mathcal{J}_{\text{TIS-DPO}}(\pi_{\text{LLM}})$, we again assume that the winning and losing examples are unambiguous, i.e. no ``losing'' response appears in another triplet as a ``winning'' response given the same prompt, and we choose $\zeta_{\text{TIS-DPO}}(y)$ as follows:
\begin{align*}
    \zeta_{\text{TIS-DPO}}(y) &\propto
    \pi_{\text{LLM}}(y|x)\prod_{i=1}^{|y|}\left(\frac{\pi_{\text{ref}}(y_i|x,y_{<i})}{\pi_{\text{LLM}}(y_i|x,y_{<i})}\right)^{w(y_i;x)}\exp\Bigg(D_{\text{SeqL}}(x,y, w(y;x); \pi_{\text{LLM}}||\pi_{\text{ref}})\Bigg),
\end{align*}
with the following token-level weight function $w(y;x) = \{w(y_i;x)\}^{|y|}_{i=1}$ defined as:
\begin{align*}
    w(y_i;x) = \begin{cases}
        w^w_i & \text{if } \ y = y_w \ \ \text{for } \ x\\
        w^l_i & \text{if } \  y = y_l \ \ \text{ for } \ x\\
    \end{cases}
\end{align*}
With $\zeta_{\text{TIS-DPO}}(y)$, we can again show another special case, whereby: 
\begin{equation*}
    \boxed{\mathcal{J}_{\text{MI-DPO}}\left(\pi_{\text{LLM}}, \textcolor{red}{\zeta \propto \zeta_{\text{TIS-DPO}}}\right)\rightarrow\mathcal{J}_{\text{TIS-DPO}}(\pi_{\text{LLM}})}
\end{equation*}

\paragraph{Recovering Sparse Preference Optimisation:} The SparsePO technique proposes a novel method for aligning large language models (LLMs) with human preferences by introducing token-level sparsity in preference optimisation \cite{christopoulou2024sparsepocontrollingpreferencealignment}. Unlike traditional approaches, SparsePO assigns weights to specific tokens during training, focusing optimisation efforts on the most relevant parts of the response. By learning sparse masks for token-level rewards and KL divergence, SparsePO improves flexibility and diversity in generated responses while maintaining alignment with human preferences. Its loss is given by: 
\begin{equation*}
    \mathcal{J}_{\text{sparsePO}}(\pi_{\text{LLM}}) = - \mathbb{E}_{(x, y_2, y_l)}\left[u_{\mu_{1}}(x,y_w, y_l) - \Delta_{\mu_2}(x,y_w,y_l) \right],
\end{equation*}
such that $u_{\mu_{1}}(x,y_w, y_l)$ is defined as: 
\begin{align*}
    u_{\mu_{1}}(x,y_w, y_l) = \alpha\sum_{t=1}^{|y_w|}\mu_1\left(y_w^{t}\right)\log\frac{\pi_{\text{LLM}}\left(\cdot|x, y_{w}^{<t}\right)}{\pi_{\text{ref}}\left(\cdot|x, y_{w}^{<t}\right)} - \alpha\sum_{t=1}^{|y_l|}\mu_1\left(y_l^{t}\right)\log\frac{\pi_{\text{LLM}}\left(\cdot|x, y_{l}^{<t}\right)}{\pi_{\text{ref}}\left(\cdot|x, y_{l}^{<t}\right)}.
\end{align*}
Furthermore, the second term $\Delta_{\mu_2}(x,y_w,y_l)$ is an altered version of TDPO's sequential Kullback-Leibler component: 
\begin{align*}
    \Delta_{\mu_2}(x, y_w, y_l) &= \alpha\sum_{t=1}^{|y_l|} \mu_2\left(y_l^{t}\right)\text{KL}\left(\pi_{\text{ref}}(\cdot|[x, y_{l}^{<t}])||\pi_{\text{LLM}}(\cdot|[x, y_{l}^{<t}])\right)  \\ 
    & \hspace{12em}- \alpha\sum_{t=1}^{|y_w|}\mu_2\left(y_w^{t}\right) \text{KL}\left(\pi_{\text{ref}}(\cdot|[x, y_{w}^{<t}])||\pi_{\text{LLM}}(\cdot|[x, y_{w}^{<t}])\right).
\end{align*}
In the above, $\mu_1(\cdot)$ and $\mu_2(\cdot)$ are masks that are applied token-level and designed to encourage sparsity. We can easily recover sparse PO using the following $\zeta$:
\begin{align*}
    \zeta_{\text{sparsePO}}(y) &\propto  \exp\Bigg(\sum_{t=1}^{|y|}\log \frac{[\pi_{\text{LLM}}(\cdot|[x,y^{<t}])]^{1-\mu_1 (y_t)}}{\pi_{\text{ref}}(.|[x,y^{<t}])^{-\mu_1(y^t)}} \\
    &\hspace{10em}-\sum_{t=1}^{|y|}\mu_2\left(y^{t}\right) \text{KL}\left(\pi_{\text{ref}}(\cdot|[x, y^{<t}])||\pi_{\text{LLM}}(\cdot|[x, y^{<t}])\right)\Bigg). 
\end{align*}
With this $\zeta_{\text{sparsePO}}(y)$ as defined above, we can show that the work in \cite{christopoulou2024sparsepocontrollingpreferencealignment} is recovered since: 
\begin{equation*}
    \boxed{\mathcal{J}_{\text{MI-DPO}}\left(\pi_{\text{LLM}}, \textcolor{red}{\zeta \propto \zeta_{\text{sparsePO}}}\right)\rightarrow\mathcal{J}_{\text{sparsePO}}(\pi_{\text{LLM}})}
\end{equation*}

\section{Discussion \& Future Work} \label{Sec:GeneralTheory}
\subsection{On Further Extensions}
We successfully recovered several special cases in the literature in the previous section. However, some still fall outside the scope of our current framework, particularly those involving global constraints. By incorporating additional constraints into our original problem formulation, we can extend our approach to recover many of these algorithms. Specifically, we can transform our original problem into a constrained one in the form of: 
\begin{align*}
    \max_{\pi_{\text{LLM}}, \zeta(y)}\sum_{x, y} p(x)\pi_{\text{LLM}}(y|x) r(x,y) - \frac{1}{\beta} \sum_{x} p(x) \text{KL}\left(\pi_{\text{LLM}}(y|x)||\zeta(y)\right), \ \ \text{s.t.} \ \ g\left(\pi_{\text{LLM}},\zeta\right) \leq \epsilon.
\end{align*}
Specifying those constraints $g(\cdot)$, would then give us the ability to unify further algorithms like CPO \cite{xu2024contrastivepreferenceoptimizationpushing} and ORPO \cite{hong2024orpomonolithicpreferenceoptimization}, among others.  

There have also been notable extensions in the literature focusing on the robustness and safety of LLMs \cite{badrinath2024hybridpreferenceoptimizationaugmenting,liang2024roporobustpreferenceoptimization,liu2024enhancingllmsafetyconstrained,ramesh2024grouprobustpreferenceoptimization}, which we would like to explore further in future work. We believe that incorporating appropriate constraints into our framework can also address these concerns. By extending our unified approach to include robustness and safety considerations, we aim to ensure that the models align with human intentions and exhibit reliable and safe behaviour across various scenarios.

While incorporating additional constraints allows us to unify a broader range of algorithms, we are hesitant to pursue this direction, as it heuristically introduces constraints without a principled foundation. Instead, we aim to explore whether these constraint-based objectives can be recovered from a probabilistic perspective in the future. We hope to derive such constraints more formally and systematically by introducing appropriate priors.

\subsection{On Optimising for $\zeta(y)$}
In the previous sections, we presented a generalised DPO loss function under general priors and explored special cases that recover well-known algorithms in the literature. We can also focus on the optimisation process over policies and priors, examining the theoretical guarantees associated with an algorithm that alternates between these two steps. The intuition behind such an approach is straightforward: instead of fixing $\zeta$, we treat it as a learnable functional that adapts alongside the policy $\pi_{\text{LLM}}$. This allows the prior $\zeta$ to act as a guide, dynamically shaping the optimisation landscape for $\pi_{\text{LLM}}$ in a manner that better aligns with the objective function. Below, we formalise this algorithm and detail its theoretical guarantees, highlighting how this approach recovers improved minima that outperform existing methods.

We can, for example, optimise $\mathcal{J}_{\text{MI-DPO}}(\pi_{\text{LLM}}, \zeta)$ jointly with respect to both arguments:
\begin{align*}
    \pi^{\star}_{\text{LLM}}, \zeta^{\star} = \arg\min_{\pi\in \Pi, \zeta\in \Xi}\mathcal{J}_{\text{MI-DPO}}\left(\pi, \zeta\right)
\end{align*}
where $\Pi$ and $\Xi$ are constraints that restrict the parameter spaces of $\pi_{\text{LLM}}$ and $\zeta$ to ensure compact domains. 

A (rather simple) realisation we notice is that jointly optimising over $\pi_{\text{LLM}}$ and $\zeta$ leads to better optimal loss values compared to any fixed $\zeta$, such that: 
\begin{equation*}
    \min_{\pi_{\text{LLM}},\zeta} \mathcal{J}_{\text{MI-DPO}}(\pi_{\text{LLM}}, \zeta) \leq \min_{\pi} \mathcal{J}_{\text{MI-DPO}}(\pi_{\text{LLM}}, \zeta) \ \ \forall \zeta.
\end{equation*}
The above result says that joint optimisation is better than optimising for $\pi_{\text{LLM}}$ alone. This is intuitive since joint optimisation gives more degrees of freedom for the loss to decrease. To see this, let $\left(\pi_{\text{LLM}}^{\star}, \zeta^{\star}\right) = \arg \min_{\pi_{\text{LLM}},\zeta}\mathcal{J}_{\text{MI-DPO}}(\pi_{\text{LLM}}, \zeta)$. This means that: 
\begin{equation}\label{joint-minimiser-def}
    \mathcal{J}_{\text{MI-DPO}}(\pi^{\star}_{\text{LLM}}, \zeta^{\star}) \leq \mathcal{J}_{\text{MI-DPO}}(\pi_{\text{LLM}}, \zeta) \ \ \forall \pi_{\text{LLM}}, \ \zeta. 
\end{equation}
Now, let us fix $\zeta = \hat{\zeta}$ and optimise over $\pi_{\text{LLM}}$ to obtain $\pi_{\text{LLM}}^{\star}(\hat{\zeta})$, such that: 
\begin{equation*}
    \pi_{\text{LLM}}^{\star}(\hat{\zeta}) = \arg\min_{\pi_{\text{LLM}}} \mathcal{J}_{\text{MI-DPO}}(\pi_{\text{LLM}},\hat{\zeta}). 
\end{equation*}
Since Equation \ref{joint-minimiser-def} holds, we conclude that $\mathcal{J}_{\text{MI-DPO}}(\pi^{\star}_{\text{LLM}},\zeta^{\star}) \le \mathcal{J}_{\text{MI-DPO}}(\pi^{\star}_{\text{LLM}}(\hat{\zeta}),\hat{\zeta})$, and, hence 
\begin{equation*}
    \boxed{ \min_{\pi_{\text{LLM}},\zeta} \mathcal{J}_{\text{MI-DPO}}(\pi_{\text{LLM}},\zeta) \leq \min_{\pi_{\text{LLM}}} \mathcal{J}_{\text{MI-DPO}}(\pi_{\text{LLM}},\hat{\zeta}) \ \ \forall \hat{\zeta}}
\end{equation*}
While this result is intuitive, it is significant because it asserts that optimising over both the policy and the prior leads to a better minimum of the loss function compared to any fixed prior. Except for DICE, the special cases we considered earlier, which recover algorithms from the literature, typically use a fixed prior. Thus, theoretically speaking, our approach of optimising over the prior will yield a better loss function value than any of these existing methods. Furthermore, when considering DICE, if the optimal policy update does not reach the optimal prior, our algorithm will still achieve better optimal loss values. 
\subsection{On Experimentation}
This paper primarily focused on theoretical advancements, providing valuable insights into the organisation of existing literature and offering a unifying framework for many DPO algorithms. From a theoretical standpoint, we also demonstrated that optimising for the priors can lead to improved loss function values. Of course, these theoretical findings must be empirically validated through experiments. Looking ahead, we plan to conduct extensive experiments to compare the general framework we propose with currently available methods in the literature, ensuring that practical results support our theoretical insights.
\break{}

\bibliographystyle{plain} 
\bibliography{main}


\newpage
\appendix
\section{Closed Form Policies}\label{App:DerivingDuals}
\begin{equation*}
    \max_{\pi_{\text{LLM}}} \mathbb{E}_{x, y \sim \pi_{\text{LLM}}(y|x)}\left[r(x,y) - \frac{1}{\beta} \log \frac{\pi_{\text{LLM}}(y|x)}{\zeta(y)}\right] = \mathcal{J}(\pi_{\text{LLM}}) \ \ \text{s.t} \ \ \sum_{y} \pi_{\text{LLM}}(y|x) = 1.
\end{equation*}
Now, let us work with the following: 
\begin{align*}
    \mathcal{J}_{{x}}(\pi_{\text{LLM}})&= \mathbb{E}_{y \sim \pi_{\text{LLM}}(y|x)}\left[r(x,y) - \frac{1}{\beta}\log \pi_{\text{LLM}}(y|x)+\frac{1}{\beta}\log\zeta(y)\right] \\
    & = \sum_{y} \pi_{\text{LLM}}(y|x)r(x,y) - \frac{1}{\beta} \sum_{y} \pi_{\text{LLM}}(y|x) \log  \pi_{\text{LLM}}(y|x) + \frac{1}{\beta} \sum_{y} \pi_{\text{LLM}}(y|x)\log \zeta(y) \\
    & = \sum_{y}  \pi_{\text{LLM}}(y|x)\left[r(x,y)+\frac{1}{\beta}\log \zeta(y)\right]+\frac{1}{\beta}\mathcal{H}\left( \pi_{\text{LLM}}(y|x)\right),
\end{align*}
with the constraint that: $\sum_{y}\pi_{\text{LLM}}(y|x) = 1$. We can, thus, define the Lagrangian:
\begin{align*}
    \mathcal{L}(\pi_{\text{LLM}}(y|x), \lambda) = \sum_{y}  \pi_{\text{LLM}}(y|x)\left[r(x,y)+\frac{1}{\beta}\log \zeta(y)\right]+\frac{1}{\beta}\mathcal{H}\left( \pi_{\text{LLM}}(y|x)\right) + \lambda(\sum_{y}\pi_{\text{LLM}}(y|x) - 1)
\end{align*}
Taking derivative with respect to $\pi_{\text{LLM}}(y|x)$:
\begin{align*}
    \frac{\partial}{\partial \pi_{\text{LLM}}(y|x)}\mathcal{L}(\pi_{\text{LLM}}(y|x), \lambda) = r(x,y) + \frac{1}{\beta} \log \zeta(y)-\frac{1}{\beta} \log \pi_{\text{LLM}}(y|x) -\frac{1}{\beta} + \lambda.
\end{align*}
Hence, for any $y$:
\begin{align*}
    &\pi^*_{\text{LLM}}(y|x) = \zeta(y)\text{exp}(\beta r(x,y) -1)\exp(\beta\lambda)
\end{align*}
Putting back to Lagrangian, we formulate dual:
\begin{align*}
    &q(\lambda) = \sum_{y}\zeta(y)\text{exp}(\beta r(x,y) -1)\exp(\beta\lambda)\left[r(x,y) + \frac{1}{\beta}\log\zeta(y)\right] - \\\nonumber
    &\frac{1}{\beta}\sum_{y}\text{exp}(\beta r(x,y) -1)\exp(\beta\lambda)\left(\beta r(x,y) + \log\zeta(y) - 1 + \lambda\beta\right) + \\\nonumber
    &\lambda\left(\sum_{y}\zeta(y)\text{exp}(\beta r(x,y) -1)\exp(\beta\lambda) - 1\right) = \\\nonumber
    &\frac{1}{\beta}\sum_{y}\zeta(y)\text{exp}(\beta r(x,y) -1)\exp(\beta\lambda) - \lambda
\end{align*}
Taking derivative for the dial with respect to $\lambda$ gives:
\begin{align*}
    \frac{d}{d\lambda}q(\lambda) = \frac{1}{\beta}\sum_{y}\zeta(y)\text{exp}(\beta r(x,y) -1)\beta\exp(\beta\lambda) - 1 = 0
\end{align*}
Hence
\begin{align*}
    \lambda^* = -\frac{1}{\beta}\log\left(\sum_{y}\zeta(y)\exp(\beta r(x,y) - 1)\right)
\end{align*}

This eventually gives:
\begin{align*}
    &\pi^*_{\text{LLM}}(y|x) = \zeta(y)\text{exp}(\beta r(x,y) -1)\exp(\beta\lambda^*) = \frac{\zeta(y)\text{exp}(\beta r(x,y) -1)}{\sum_{y}\zeta(y)\text{exp}(\beta r(x,y) -1)}.
\end{align*}
\subsection{Deriving the Generalised Mutual Information DPO's Loss}
We can follow the standard approach presented in \cite{rafailov2024direct} to derive our new DPO loss. Here, we will start from the Bradley-Terry model and assume that the probability an answer $y_1$ is preferred to an answer $y_2$ is given by: 
\begin{align*}
    \textbf{Pr}(y_1 \succ y_2|x) &= \frac{\exp(r(x,y_1))}{\exp(r(x,y_1)) + \exp(r(x,y_2))} 
\end{align*}
From the derivation of the optimal policy under mutual information reguralisation we found that: 
\begin{equation*}
    \pi_{\text{LLM}}^{\star}(y|x) = \frac{1}{\hat{\mathcal{Z}}} \zeta(y) \exp(\beta r(x,y)). 
\end{equation*}
Therefore, we derive our reward function as: 
\begin{align*}
\exp(\beta r(x,y)) &= \hat{\mathcal{Z}}\frac{\pi_{\text{LLM}}^{\star}(y|x)}{\zeta(y)} \implies r(x,y) = \frac{1}{\beta}\log\left[\hat{\mathcal{Z}}\frac{\pi_{\text{LLM}}^{\star}(y|x)}{\zeta(y)}\right]   \\
 r(x,y) &= \frac{1}{\beta}\left[\log \pi_{\text{LLM}}^{\star}(y|x) + \log \hat{\mathcal{Z}} - \log \zeta(y)\right] \\
 & = \frac{1}{\beta} \log \frac{\pi_{\text{LLM}}^{\star}(y|x)}{\zeta(y)} + \frac{1}{\beta} \log \hat{\mathcal{Z}} 
\end{align*}
Now, let us substitute those into the Bradley-Terry's model: 
\begin{align*}
\textbf{Pr}(y_1 \succ y_2|x) &= \frac{\exp\left(\frac{1}{\beta} \log \frac{\pi_{\text{LLM}}^{\star}(y_{1}|x)}{\zeta(y_{1})} + \frac{1}{\beta} \log \hat{\mathcal{Z}}\right)}{\exp\left(\frac{1}{\beta} \log \frac{\pi_{\text{LLM}}^{\star}(y_1|x)}{\zeta(y_1)} + \frac{1}{\beta} \log \hat{\mathcal{Z}}\right) + \exp\left(\frac{1}{\beta} \log \frac{\pi_{\text{LLM}}^{\star}(y_2|x)}{\zeta(y_2)} + \frac{1}{\beta} \log \hat{\mathcal{Z}}\right)} \\
& = \frac{\exp\left(\frac{1}{\beta} \log \hat{\mathcal{Z}}\right) \exp\left(\frac{1}{\beta}\log\frac{\pi_{\text{LLM}}^{\star}(y_1|x)}{\zeta(y_1)}\right)}{\exp\left(\frac{1}{\beta} \log \hat{\mathcal{Z}}\right)\left[\exp\left(\frac{1}{\beta}\log\frac{\pi_{\text{LLM}}^{\star}(y_1|x)}{\zeta(y_1)}\right)+\exp\left(\frac{1}{\beta}\log\frac{\pi_{\text{LLM}}^{\star}(y_2|x)}{\zeta(y_2)}\right)\right]} \\
 & = \text{sigmoid}\left(\alpha \log \frac{\pi_{\text{LLM}}^{\star}(y_1|x)}{\zeta(y_1) } - \alpha \log \frac{\pi_{\text{LLM}}^{\star}(y_2|x)}{\zeta(y_2) } \right),
\end{align*}
with $\alpha = \frac{1}{\beta}$.

Taking the logarithm of the above gives our new loss: 
\begin{equation*}
    \mathcal{J}_{\text{MI-DPO}}\left(\pi_{\text{LLM}}, \zeta\right) = - \mathbb{E}_{(x, y_w, y_l) \sim \mathcal{D}}\left[\log \text{sigmoid}\left(\alpha \log \frac{\pi_{\text{LLM}}(y_w|x)}{\zeta(y_w) } - \alpha \log \frac{\pi_{\text{LLM}}(y_l|x)}{\zeta(y_l) } \right) \right].
\end{equation*}
\section{Special Case Choices of $\zeta(y)$}\label{App:SpecialCases}
This section demonstrates how current variants of DPO can be seen as a special case of $ \mathcal{J}_{\text{MI-DPO}}$ when choosing the correct prior $\zeta(y)$. 

\paragraph{Recovering Standard DPO:} It is easy to see that if we choose $\alpha = 1$ and $\zeta(y) = \pi_{\text{ref}}(y|x)$, our equation boils down to standard DPO giving us: 
\begin{equation*}
    \boxed{\mathcal{J}_{\text{MI-DPO}}\left(\pi_{\text{LLM}}, \textcolor{red}{\zeta\equiv\pi_{\text{ref}}}\right) \rightarrow \mathcal{J}_{\text{DPO}}}
\end{equation*}
\paragraph{Recovering DICE \cite{chen2024bootstrappinglanguagemodelsdpo}:} At any iteration $t$, the loss function of DICE is given by:
\begin{equation*}
    \mathcal{J}_{\text{DICE}}(\pi_{\text{LLM}}) = - \mathbb{E}_{(x, y_w, y_l) \sim \mathcal{D}_t}\left[\log \text{sigmoid}\left(\alpha \log \frac{\pi_{\text{LLM}}(y_w|x)}{\pi^{(t-1)}_{\text{LLM}}(y_w|x) } - \alpha \log \frac{\pi_{\text{LLM}}(y_l|x)}{\pi^{(t-1)}_{\text{LLM}}(y_l|x) } \right) \right],
\end{equation*}
where $\pi^{(t-1)}_{\text{LLM}}(y|x)$ is the previous iteration updated policy. Assuming we apply MI-DPO in an iterative fashion like DICE, we can easily see that setting $\zeta$ to $\zeta_{\text{DICE}}(y) = \pi_{\text{LLM}}^{(t-1)}\left(y|x\right)$, we get: 
\begin{equation*}
    \boxed{
    \mathcal{J}_{\text{MI-DPO}}\left(\pi_{\text{LLM}}, \textcolor{red}{\zeta \equiv \zeta_{\text{DICE}}}\right)
     \rightarrow \mathcal{J}_{\text{DICE}}(\pi_{\text{LLM}}). }
\end{equation*}
\paragraph{Recovering Entropy Controllable DPO \cite{omura2024entropycontrollabledirectpreference}:} Entropy controllable DPO is a simple extension to standard DPO incorporating another parameter $\lambda$, allowing for entropy adjustment. Its loss is defined by: 
\begin{equation*}
    \mathcal{J}_{\text{cEntopy}}(\pi_{\text{LLM}}) = - \mathbb{E}_{(x, y_w, y_l) \sim \mathcal{D}_t}\left[\log \text{sigmoid}\left(\alpha \log \frac{\pi^{\lambda}_{\text{LLM}}(y_w|x)}{\pi_{\text{ref}}(y_w|x) } - \alpha \log \frac{\pi^{\lambda}_{\text{LLM}}(y_l|x)}{\pi_{\text{ref}}(y_l|x) } \right) \right].
\end{equation*}
Setting $\zeta$ to $\zeta_{\text{cEntropy}}(y) = \pi_{\text{ref}}(y|x)[\pi_{\text{LLM}}(y|x)]^{1-\lambda}$, we easily see that: 
\begin{equation*}
   \boxed{ \mathcal{J}_{\text{MI-DPO}}(\pi_{\text{LLM}}, \textcolor{red}{\zeta \propto \zeta_{\text{cEntropy}}}) \rightarrow \mathcal{J}_{\text{cEntropy}}(\pi_{\text{LLM}}). }
\end{equation*}
\paragraph{Recovering SimPO \cite{meng2024simposimplepreferenceoptimization}:} The standard loss for SimPO is given by: 
\begin{equation*}
    \mathcal{J}_{\text{SimPO}} (\pi_{\text{LLM}}) = - \mathbb{E}_{(x, y_w, y_l) \sim \mathcal{D}}\left[\log \text{sigmoid}\left(\frac{\alpha}{|y_w|}\log \pi_{\text{LLM}}(y_w|x) - \frac{\alpha}{|y_l|}\log \pi_{\text{LLM}}(y_l|x)-\gamma\right)\right],  
\end{equation*}
where $|y|$ denotes the length of an answer $y$, and $\alpha$ and $\gamma$ are constants. 

We can recover $\mathcal{J}_{\text{simPO}}$ from $\mathcal{J}_{\text{MI-DPO}}$ by choosing the following for the prior: 

\begin{equation*}
    \zeta_{\text{simPO}}(y) \propto \left[\pi_{\text{LLM}}(y|x)\right]^{\left(1-\frac{1}{|y|}\right)}\exp{\beta(y,x)}\ \ \text{with} \ \ 
    \beta(y,x) = \begin{cases}
        \frac{\gamma}{2\alpha} & \text{if } \ y = y_w \ \ \text{for } \ x,\\
        -\frac{\gamma}{2\alpha} & \text{if } \  y = y_l \ \ \text{ for } \ x.\\
    \end{cases}.
\end{equation*}


In doing so, we notice that the term of the ratio inside $\mathcal{J}_{\text{MI-DPO}}$ boils down to: 
\begin{align*}
    \alpha \log \frac{\pi_{\text{LLM}}(y|x)}{\zeta(y)} &= \alpha \log \frac{\pi_{\text{LLM}}(y|x)}{\left[\pi_{\text{LLM}}(y|x)\right]^{\left(1-\frac{1}{|y|}\right)}\exp{\beta(y,x)}} \\
    & = \alpha \log \frac{\pi_{\text{LLM}}(y|x)}{\pi_{\text{LLM}}(y|x) \pi_{\text{LLM}}(y|x)^{-\frac{1}{|y|}}\exp{\beta(y,x)}} \\
    & = \alpha \log \pi_{\text{LLM}}(y|x)^{\frac{1}{|y|}} - \alpha\beta(y,x) = \frac{\alpha}{|y|}\log \pi_{\text{LLM}}(y|x)- \alpha\beta(y,x).
\end{align*}
Therefore, we can now write that: 
\begin{align*}
    \alpha \log \frac{\pi_{\text{LLM}}(y_w|x)}{\zeta(y_w)} - \alpha \log \frac{\pi_{\text{LLM}}(y_l|x)}{\zeta(y_l)} &= \frac{\alpha}{|y_w|}\log \pi_{\text{LLM}}(y_w|x)- \alpha\beta(y_w,x) \\
    &\hspace{7em}- \frac{\alpha}{|y_l|}\log \pi_{\text{LLM}}(y_l|x)- \alpha\beta(y_l,x) \\
    & = \frac{\alpha}{|y_w|}\log \pi_{\text{LLM}}(y_w|x) - \frac{\alpha}{|y_l|}\log \pi_{\text{LLM}}(y_l|x) -\gamma
\end{align*}
Therefore, we can now see that:
\begin{equation*}
\boxed{    \mathcal{J}_{\text{MI-DPO}}\left(\pi_{\text{LLM}}(y|x), \textcolor{red}{\zeta \propto \left[\pi_{\text{LLM}}(y|x)\right]^{\left(1-\frac{1}{|y|}\right)}\exp{\beta(y,x)}}\right) \rightarrow \mathcal{J}_{\text{simPO}}(\pi_{\text{LLM}}).} 
\end{equation*}
With those results, we can say that upon choosing the correct $\zeta$ we recover the loss of simPO from \cite{meng2024simposimplepreferenceoptimization}, making it a special case of our framework. 
\paragraph{Recovering R-DPO \cite{park2024disentanglinglengthqualitydirect}:} The second algorithm we consider is R-DPO, which has a loss function defined by: 
\begin{align*}
    \mathcal{J}_{\text{R-DPO}}(\pi_{\text{LLM}}) &=-\mathbb{E}_{(x,y_w, y_l) \sim \mathcal{D}} \Bigg[\log \text{sigmoid}\Bigg(\alpha \log \frac{\pi_{\text{LLM}}(y_w|x)}{\pi_{\text{ref}}(y_w|x)} - \alpha \log \frac{\pi_{\text{LLM}}(y_l|x)}{\pi_{\text{ref}}(y_l|x)} \\ 
    &\hspace{25em} + \underbrace{\Bigg(\tilde{\alpha} |y_w| - \tilde{\alpha} |y_l|\Bigg)}_{\text{regulariser}}
    \Bigg)\Bigg].
\end{align*}
To recover $ \mathcal{J}_{\text{R-DPO}}(\pi_{\text{LLM}})$ from $ \mathcal{J}_{\text{MI-DPO}}(\pi_{\text{LLM}})$, we simply need to set $\zeta(y)$ as follows: 
\begin{equation*}
    \zeta_{\text{R-DPO}}(y) \propto \frac{\pi_{\text{ref}}(y|x)}{\exp(\beta |y|)}.
\end{equation*}
Let us illustrate why $\zeta_{\text{R-DPO}}(y)$ would recover $ \mathcal{J}_{\text{R-DPO}}$. We start by looking at the terms inside the sigmoid of $ \mathcal{J}_{\text{MI-DPO}}$ and write for any $y$: 
\begin{align*}
    \alpha \log \frac{\pi_{\text{LLM}}(y|x)}{\zeta(y)} &= \alpha \log \frac{\pi_{\text{LLM}}(y|x)}{ \frac{\pi_{\text{ref}}(y|x)}{\exp(\beta |y|)}} \ \ \ (\text{Replacing $\zeta(y)$ with $\zeta_{\text{R-DPO}}(y)$}) \\
    & = \alpha \log \frac{\exp(\beta |y|) \pi_{\text{LLM}}(y|x)}{\pi_{\text{ref}}(y|x)} \\ 
    & = \alpha \log \frac{\pi_{\text{LLM}}(y|x)}{\pi_{\text{ref}}(y|x)} + \alpha \log \exp\left(\beta |y|\right) = \alpha \log \frac{\pi_{\text{LLM}}(y|x)}{\pi_{\text{ref}}(y|x)} + \tilde{\alpha} |y| \ \ \text{with $\tilde{\alpha} = \alpha\beta$.}
\end{align*}
Therefore, we can now write that: 
\begin{align*}
    \alpha \log \frac{\pi_{\text{LLM}}(y_w|x)}{\zeta(y_w)} - \alpha \log \frac{\pi_{\text{LLM}}(y_l|x)}{\zeta(y_l)} &= \alpha \log \frac{\pi_{\text{LLM}}(y_w|x)}{\pi_{\text{ref}}(y_w|x)} + \tilde{\alpha} |y_w|  - \alpha \log \frac{\pi_{\text{LLM}}(y_l|x)}{\pi_{\text{ref}}(y_l|x)} - \tilde{\alpha} |y_l| \\
    & = \alpha \log \frac{\pi_{\text{LLM}}(y_w|x)}{\pi_{\text{ref}}(y_w|x)} - \alpha \log \frac{\pi_{\text{LLM}}(y_l|x)}{\pi_{\text{ref}}(y_l|x)} + \tilde{\alpha}(|y_w| - |y_l|).
\end{align*}
The last term in the above equation is exactly the inner term inside the sigmoid of $\mathcal{J}_{\text{R-DPO}}$. Hence, we conclude another special case in that: 
\begin{equation*}
\boxed{
    \mathcal{J}_{\text{MI-DPO}}\left(\pi_{\text{LLM}}(y|x), \textcolor{red}{\zeta \propto \frac{\pi_{\text{ref}}(y|x)}{\exp(\beta |y|)}}\right) \rightarrow \mathcal{J}_{\text{R-DPO}}.} 
\end{equation*}
\paragraph{Recovering TDPO \cite{zeng2024tokenleveldirectpreferenceoptimization}:} When it comes to TDPO, the loss function is given by: 
\begin{align*}
    \mathcal{J}_{\text{TDPO}}(\pi_{\text{LLM}}) &= -\mathbb{E}_{(x, y_w, y_l)\sim \mathcal{D}}\Bigg[\log \text{sigmoid}\Bigg(\alpha \log \frac{\pi_{\text{LLM}}(y_w|x)}{\pi_{\text{ref}}(y_w|x)} - \alpha \log \frac{\pi_{\text{LLM}}(y_l|x)}{\pi_{\text{ref}}(y_l|x)} \\ 
   & \hspace{27em} \underbrace{-\alpha\Delta(x, y_w, y_l)}_{\text{regulariser}}\Bigg)\Bigg], 
\end{align*}
with $\Delta(x, y_w, y_l)$ defined as follows: 
\begin{equation*}
    \Delta(x, y_w, y_l) = \sum_{t=1}^{|y_l|} \text{KL}\left(\pi_{\text{ref}}(\cdot|[x, y_{l}^{<t}])||\pi_{\text{LLM}}(\cdot|[x, y_{l}^{<t}])\right) - \sum_{t=1}^{|y_w|} \text{KL}\left(\pi_{\text{ref}}(\cdot|[x, y_{w}^{<t}])||\pi_{\text{LLM}}(\cdot|[x, y_{w}^{<t}])\right).
\end{equation*}
Now, if we are to set our prior as follows: 
\begin{equation*}
    \zeta_{\text{TDPO}}(y) \propto \pi_{\text{ref}}(y|x) \exp\left(-\sum_{t=1}^{|y|}\text{KL}\left(\pi_{\text{ref}}\left(\cdot|x, y^{<t}\right)||\pi_{\text{LLM}}\left(\cdot|x, y^{<t}\right) \right)\right).
\end{equation*}
With this choice, we can recover $\mathcal{J}_{\text{TDPO}}$ as yet another special case of our formulation. To see this fact, let us evaluate the following term from $\mathcal{J}_{\text{MI-DPO}}$: 
\begin{align*}
    \alpha\log \frac{\pi_{\text{LLM}}(y|x)}{\zeta(y)} \ \ \text{with $\zeta(y) = \zeta_{\text{TDPO}}(y)$}. 
\end{align*}
This leads us to: 
\begin{align*}
   \alpha\log \frac{\pi_{\text{LLM}}(y|x)}{\zeta_{\text{TDPO}}(y)} &= \alpha \log \frac{\pi_{\text{LLM}}(y|x)}{\pi_{\text{ref}}(y|x) \exp\left(-\sum_{t=1}^{|y|}\text{KL}\left(\pi_{\text{ref}}\left(\cdot|x, y^{<t}\right)||\pi_{\text{LLM}}\left(\cdot|x, y^{<t}\right) \right)\right)} \\
   & = \alpha \log \frac{\pi_{\text{LLM}}(y|x)}{\pi_{\text{ref}}(y|x)} + \alpha \log \left[\frac{1}{\exp\left(-\sum_{t=1}^{|y|}\text{KL}\left(\pi_{\text{ref}}\left(\cdot|x, y^{<t}\right)||\pi_{\text{LLM}}\left(\cdot|x, y^{<t}\right) \right)\right)}\right] \\
   & = \alpha \log \frac{\pi_{\text{LLM}}(y|x)}{\pi_{\text{ref}}(y|x)} + \alpha \sum_{t=1}^{|y|}\text{KL}\left(\pi_{\text{ref}}\left(\cdot|x, y^{<t}\right)||\pi_{\text{LLM}}\left(\cdot|x, y^{<t}\right) \right).
\end{align*}
Therefore, given win and lose answers, we write: 
\begin{align*}
    \alpha\log \frac{\pi_{\text{LLM}}(y_w|x)}{\zeta_{\text{TDPO}}(y_w)} - \alpha\log \frac{\pi_{\text{LLM}}(y_l|x)}{\zeta_{\text{TDPO}}(y_l)} = \alpha \log \frac{\pi_{\text{LLM}}(y_w|x)}{\pi_{\text{ref}}(y_w|x)} - \alpha \log \frac{\pi_{\text{LLM}}(y_l|x)}{\pi_{\text{ref}}(y_l|x)} - \alpha \Delta(x,y_w,y_l),
\end{align*}
with $\Delta$ defined as: 
\begin{equation*}
    \Delta(x, y_w, y_l) = \sum_{t=1}^{|y_l|}\text{KL}\left(\pi_{\text{ref}}\left(\cdot|x, y_{l}^{<t}\right)||\pi_{\text{LLM}}\left(\cdot|x, y_{l}^{<t}\right) \right) - \sum_{t=1}^{|y_w|}\text{KL}\left(\pi_{\text{ref}}\left(\cdot|x, y_{w}^{<t}\right)||\pi_{\text{LLM}}\left(\cdot|x, y_{w}^{<t}\right) \right).
\end{equation*}
In short, upon choosing $\zeta(y)$ to be $\zeta_{\text{TDPO}}$, we recover $\mathcal{J}_{\text{TDPO}}(\pi_{\text{LLM}})$: 
\begin{equation*}
   \boxed{ \mathcal{J}_{\text{MI-DPO}}\left(\cdot, \textcolor{red}{\zeta \propto \pi_{\text{ref}}(y|x) \exp\left(-\sum_{t=1}^{|y|}\text{KL}\left(\pi_{\text{ref}}\left(\cdot|x, y^{<t}\right)||\pi_{\text{LLM}}\left(\cdot|x, y^{<t}\right) \right)\right)}\right) \rightarrow \mathcal{J}_{\text{TDPO}}\left(\cdot\right).} 
\end{equation*}
\paragraph{Recovering SparsePO \cite{christopoulou2024sparsepocontrollingpreferencealignment}:} SprasePO is an extension of TDPO to introduce sparsity masks leading to the following loss function: 
\begin{equation*}
    \mathcal{J}_{\text{sparsePO}}(\pi_{\text{LLM}}) = - \mathbb{E}_{(x, y_2, y_l)}\left[u_{\mu_{1}}(x,y_w, y_l) - \Delta_{\mu_2}(x,y_w,y_l) \right],
\end{equation*}
such that $u_{\mu_{1}}(x,y_w, y_l)$ is defined as: 
\begin{align*}
    u_{\mu_{1}}(x,y_w, y_l) = \alpha\sum_{t=1}^{|y_w|}\mu_1\left(y_w^{t}\right)\log\frac{\pi_{\text{LLM}}\left(\cdot|x, y_{w}^{<t}\right)}{\pi_{\text{ref}}\left(\cdot|x, y_{w}^{<t}\right)} - \alpha\sum_{t=1}^{|y_l|}\mu_1\left(y_l^{t}\right)\log\frac{\pi_{\text{LLM}}\left(\cdot|x, y_{l}^{<t}\right)}{\pi_{\text{ref}}\left(\cdot|x, y_{l}^{<t}\right)}.
\end{align*}
Furthermore, the second term $\Delta_{\mu_2}(x,y_w,y_l)$ is an altered version of TDPO's sequential Kullback-Leibler component: 
\begin{align*}
    \Delta_{\mu_2}(x, y_w, y_l) &= \alpha\sum_{t=1}^{|y_l|} \mu_2\left(y_l^{t}\right)\text{KL}\left(\pi_{\text{ref}}(\cdot|[x, y_{l}^{<t}])||\pi_{\text{LLM}}(\cdot|[x, y_{l}^{<t}])\right)  \\ 
    & \hspace{12em}- \alpha\sum_{t=1}^{|y_w|}\mu_2\left(y_w^{t}\right) \text{KL}\left(\pi_{\text{ref}}(\cdot|[x, y_{w}^{<t}])||\pi_{\text{LLM}}(\cdot|[x, y_{w}^{<t}])\right).
\end{align*}
In the above, $\mu_1(\cdot)$ and $\mu_2(\cdot)$ are masks that are applied token-level and designed to encourage sparsity. 
 
To recover $\mathcal{J}_{\text{sparsePO}}(\pi_{\text{LLM}})$, we choose $\zeta_{\text{sparsePO}}(y)$ as follows: 
\begin{align*}
    \zeta_{\text{sparsePO}}(y) &\propto \exp\Bigg(\log \pi_{\text{LLM}}(y|x) - \sum_{t=1}^{|y|}\mu_1\left(y^{t}\right)\log\frac{\pi_{\text{LLM}}\left(\cdot|[x, y^{<t}]\right)}{\pi_{\text{ref}}\left(\cdot|[x, y^{<t}]\right)} \\
    &\hspace{10em}-\sum_{t=1}^{|y|}\mu_2\left(y^{t}\right) \text{KL}\left(\pi_{\text{ref}}(\cdot|[x, y^{<t}])||\pi_{\text{LLM}}(\cdot|[x, y^{<t}])\right)\Bigg) \\
    & = \exp\Bigg(\sum_{t=1}^{|y|}\log \pi_{\text{LLM}}(\cdot|[x, y^{<t}]) - \sum_{t=1}^{|y|}\mu_1\left(y^{t}\right)\log\frac{\pi_{\text{LLM}}\left(\cdot|[x, y^{<t}]\right)}{\pi_{\text{ref}}\left(\cdot|[x, y^{<t}]\right)} \\
    &\hspace{10em}-\sum_{t=1}^{|y|}\mu_2\left(y^{t}\right) \text{KL}\left(\pi_{\text{ref}}(\cdot|[x, y^{<t}])||\pi_{\text{LLM}}(\cdot|[x, y^{<t}])\right)\Bigg)\\
    & =\exp\Bigg(\sum_{t=1}^{|y|}\log \pi_{\text{LLM}}(\cdot|[x, y^{<t}]) -\mu_1\left(y^{t}\right)\log\frac{\pi_{\text{LLM}}\left(\cdot|[x, y^{<t}]\right)}{\pi_{\text{ref}}\left(\cdot|[x, y^{<t}]\right)} \\
    &\hspace{10em}-\sum_{t=1}^{|y|}\mu_2\left(y^{t}\right) \text{KL}\left(\pi_{\text{ref}}(\cdot|[x, y^{<t}])||\pi_{\text{LLM}}(\cdot|[x, y^{<t}])\right)\Bigg) \\ 
    & = \exp\Bigg(\sum_{t=1}^{|y|}\log \frac{[\pi_{\text{LLM}}(\cdot|[x,y^{<t}])]^{1-\mu_1 (y_t)}}{\pi_{\text{ref}}(.|[x,y^{<t}])^{-\mu_1(y^t)}} \\
    &\hspace{10em}-\sum_{t=1}^{|y|}\mu_2\left(y^{t}\right) \text{KL}\left(\pi_{\text{ref}}(\cdot|[x, y^{<t}])||\pi_{\text{LLM}}(\cdot|[x, y^{<t}])\right)\Bigg) \\ 
\end{align*}
With $\zeta_{\text{sparsePO}}(y)$, we show that: 
\begin{align*}
    \alpha \log \frac{\pi_{\text{LLM}}(y|x)}{\zeta_{\text{sparsePO}}(y)} &= \alpha \log {\pi_{\text{LLM}}(y|x)} - \alpha \log \zeta_{\text{sparsePO}}(y) \\
    & = \alpha \log {\pi_{\text{LLM}}(y|x)} -\alpha \log \pi_{\text{LLM}}(y|x) + \alpha \sum_{t=1}^{|y|}\mu_1\left(y^{t}\right)\log\frac{\pi_{\text{LLM}}\left(\cdot|[x, y^{<t}]\right)}{\pi_{\text{ref}}\left(\cdot|[x, y^{<t}]\right)}\\ 
    & \hspace{10em} + \alpha \sum_{t=1}^{|y|}\mu_2\left(y^{t}\right) \text{KL}\left(\pi_{\text{ref}}(\cdot|[x, y^{<t}])||\pi_{\text{LLM}}(\cdot|[x, y^{<t}])\right). \\
    &  = \alpha \sum_{t=1}^{|y|}\mu_1\left(y^{t}\right)\log\frac{\pi_{\text{LLM}}\left(\cdot|[x, y^{<t}]\right)}{\pi_{\text{ref}}\left(\cdot|[x, y^{<t}]\right)}\\ 
    & \hspace{10em} + \alpha \sum_{t=1}^{|y|}\mu_2\left(y^{t}\right) \text{KL}\left(\pi_{\text{ref}}(\cdot|[x, y^{<t}])||\pi_{\text{LLM}}(\cdot|[x, y^{<t}])\right).
\end{align*}
Hence, the term inside of the sigmoid in $\mathcal{J}_{\text{MI-DPO}}$ boils down to: 
\begin{align*}
     \alpha \log \frac{\pi_{\text{LLM}}(y_w|x)}{\zeta_{\text{sparsePO}}(y_w)} -  \alpha \log \frac{\pi_{\text{LLM}}(y_l|x)}{\zeta_{\text{sparsePO}}(y_l)} & = \alpha \sum_{t=1}^{|y|}\mu_1\left(y_w^{t}\right)\log\frac{\pi_{\text{LLM}}\left(\cdot|[x, y_w^{<t}]\right)}{\pi_{\text{ref}}\left(\cdot|[x, y_w^{<t}]\right)} \\ 
     &\hspace{2em}- \alpha \sum_{t=1}^{|y|}\mu_1\left(y_l^{t}\right)\log\frac{\pi_{\text{LLM}}\left(\cdot|[x, y_l^{<t}]\right)}{\pi_{\text{ref}}\left(\cdot|[x, y_l^{<t}]\right)} - \Delta_{\mu_2}(x, y_w, y_l),
\end{align*}
with $\Delta_{\mu_2}(x, y_w, y_l)$ being defined as: 

    \begin{align*}
    \Delta_{\mu_2}(x, y_w, y_l) &= \alpha\sum_{t=1}^{|y_l|} \mu_2\left(y_l^{t}\right)\text{KL}\left(\pi_{\text{ref}}(\cdot|[x, y_{l}^{<t}])||\pi_{\text{LLM}}(\cdot|[x, y_{l}^{<t}])\right)  \\ 
    & \hspace{12em}- \alpha\sum_{t=1}^{|y_w|}\mu_2\left(y_w^{t}\right) \text{KL}\left(\pi_{\text{ref}}(\cdot|[x, y_{w}^{<t}])||\pi_{\text{LLM}}(\cdot|[x, y_{w}^{<t}])\right).
\end{align*}
Clearly, the above recovers $\mathcal{J}_{\text{sparsePO}}$, leading us to say: 
\begin{equation*}
    \boxed{\mathcal{J}_{\text{MI-DPO}}\left(\pi_{\text{LLM}}, \textcolor{red}{\zeta \propto \zeta_{\text{sparsePO}}}\right)\rightarrow\mathcal{J}_{\text{sparsePO}}(\pi_{\text{LLM}}).}
\end{equation*}
\paragraph{Recovering TIS-DPO \cite{liu2024tisdpotokenlevelimportancesampling}:} 
TIS-DPO is defined via loss function: 
\begin{equation*}
    \mathcal{J}_{\text{TIS-DPO}} = -\mathbb{E}_{(x,y_w,y_l)\sim\mathcal{D}}\left[\log\sigma\left(u(x,y_w,y_{l},\pi_{\text{LLM}}, w^w, w^l ) - \eta(x,y_w,y_{l},\pi_{\text{LLM}}, w^w, w^l )\right)\right],
\end{equation*}
such that $u(x,y_w,y_{l},\pi_{\text{LLM}}, w^w, w^l )$ is defined as: 
\begin{align*}
    u(x,y_w,y_{l},\pi_{\text{LLM}}, w^w, w^l ) = \sum_{i=1}^{T_w}w^w_i\alpha\log\frac{\pi_{LLM}(y_{w_i}|x, y_{w_{<i}})}{\pi_{\text{ref}}(y_{w_i}|x,y_{w_{<i}})} - \sum_{j=1}^{T_l}w^l_j\alpha\log\frac{\pi_{LLM}(y_{l_j}|x, y_{l_{<j}})}{\pi_{\text{ref}}(y_{l_j}|x,y_{l_{<j}})}.
\end{align*}
Furthermore, the second term $\eta(x,y_w,y_{l},\pi_{\text{LLM}}, w^w, w^l )$ is an altered version of TDPO's sequential Kullback-Leibler component: 
\begin{align*}
    \eta(x,y_w,y_{l},\pi_{\text{LLM}}, w^w, w^l ) = \alpha D_{\text{SeqL}}(x,y_w, w^w; \pi_{\text{LLM}}||\pi_{\text{ref}}) - \alpha D_{\text{SeqL}}(x,y_l, w^l; \pi_{\text{LLM}}||\pi_{\text{ref}}).
\end{align*}
In the above, $w^w_i$ and $w^l_j$ are weights  that are applied token-level. Finally, $D_{\text{SeqL}}(x,y_l, w^l; \pi_{\text{LLM}}||\pi_{\text{ref}})$ is weighted sequence KL-divergence defined as:
\begin{align*}
    &D_{\text{SeqL}}(x,y, \tilde{w}; \pi_{1}||\pi_{2}) = \sum_{i=1}^{|y|}\tilde{w}_i\text{KL}(\pi_1(y_i| x, y_{<i}) || \pi_2(y_i| x, y_{<i}))
\end{align*}
To recover $\mathcal{J}_{\text{TIS-DPO}}(\pi_{\text{LLM}})$, we choose $\zeta_{\text{TIS-DPO}}(y)$ as follows:
\begin{align*}
    \zeta_{\text{TIS-DPO}}(y) &\propto
    \pi_{\text{LLM}}(y|x)\prod_{i=1}^{|y|}\left(\frac{\pi_{\text{ref}}(y_i|x,y_{<i})}{\pi_{\text{LLM}}(y_i|x,y_{<i})}\right)^{w(y_i;x)}\exp\Bigg(D_{\text{SeqL}}(x,y, w(y;x); \pi_{\text{LLM}}||\pi_{\text{ref}})\Bigg)\\
    & = \prod_{i=1}^{|y|}\frac{\left(\pi_{\text{ref}}(y_i|x,y_{<i})\right)^{w(y_i;x)}}{\left(\pi_{\text{LLM}}(y_i|x,y_{<i})\right)^{w(y_i;x)-1}}\exp\Bigg(\sum_{i=1}^{|y|}w(y_i;x)\text{KL}(\pi_{\text{LLM}}(y_i| x, y_{<i}) || \pi_{\text{ref}}(y_i| x, y_{<i}))\Bigg)
\end{align*}
with token-level weight function $w(y;x) = \{w(y_i;x)\}^{|y|}_{i=1}$ defined as:
\begin{align*}
    w(y_i;x) = \begin{cases}
        w^w_i & \text{if } \ y = y_w \ \ \text{for } \ x\\
        w^l_i & \text{if } \  y = y_l \ \ \text{ for } \ x\\
    \end{cases}
\end{align*}

With $\zeta_{\text{TIS-DPO}}(y)$, we show that: 
\begin{align*}
    \alpha \log \frac{\pi_{\text{LLM}}(y|x)}{\zeta_{\text{TIS-DPO}}(y)} &= -\alpha\log \left[\prod_{i=1}^{|y|}\left(\frac{\pi_{\text{ref}}(y_i|x,y_{<i})}{\pi_{\text{LLM}}(y_i|x,y_{<i})}\right)^{w(y_i;x)}\exp\Bigg(D_{\text{SeqL}}(x,y, w(y;x); \pi_{\text{LLM}}||\pi_{\text{ref}})\Bigg)\right]\\
    & = -\alpha D_{\text{SeqL}}(x,y, w(y;x); \pi_{\text{LLM}}||\pi_{\text{ref}})  + \alpha \sum_{i=1}^{|y|}w(y_i;x)\log\frac{\pi_{\text{LLM}}\left(y_i|x, y_{<i}\right)}{\pi_{\text{ref}}\left(y_i|x, y_{<i}\right)}.
\end{align*}
Hence, the term inside of the sigmoid in $\mathcal{J}_{\text{MI-DPO}}$ boils down to: 
\begin{align*}
     \alpha \log \frac{\pi_{\text{LLM}}(y_w|x)}{\zeta_{\text{TIS-DPO}}(y_w)} -  \alpha \log \frac{\pi_{\text{LLM}}(y_l|x)}{\zeta_{\text{TIS-DPO}}(y_l)} & = \alpha \sum_{i=1}^{|y_w|}w(y_{w_i};x)\log\frac{\pi_{\text{LLM}}\left(y_{w_i}|x, y_{w_{<i}}\right)}{\pi_{\text{ref}}\left(y_{w_i}|x, y_{w_{<i}}\right)} \\ 
     &\hspace{2em}- \alpha\sum_{j=1}^{|y_l|}w(y_{l_j};x)\log\frac{\pi_{\text{LLM}}\left(y_{l_j}|x, y_{l_{<j}}\right)}{\pi_{\text{ref}}\left(y_{l_j}|x, y_{l_{<j}}\right)}\\
      &\hspace{2em}-\alpha D_{\text{SeqL}}(x,y_w, w(y_w;x); \pi_{\text{LLM}}||\pi_{\text{ref}})\\
      &\hspace{2em}+\alpha D_{\text{SeqL}}(x,y_l, w(y_l;x); \pi_{\text{LLM}}||\pi_{\text{ref}}).
\end{align*}
applying $w(y_{w_i};x) = w^w_i, \ \ w(y_{l_j};x) = w^l_j$ and expressions for $u(x,y_w,y_{l},\pi_{\text{LLM}}, w^w, w^l )$ and $\eta(x,y_w,y_{l},\pi_{\text{LLM}}, w^w, w^l )$ gives:
\begin{align*}
     \alpha \log \frac{\pi_{\text{LLM}}(y_w|x)}{\zeta_{\text{TIS-DPO}}(y_w)} -  \alpha \log \frac{\pi_{\text{LLM}}(y_l|x)}{\zeta_{\text{TIS-DPO}}(y_l)} & = u(x,y_w,y_{l},\pi_{\text{LLM}}, w^w, w^l ) \\ 
     &\hspace{2em}- \eta(x,y_w,y_{l},\pi_{\text{LLM}}, w^w, w^l ).
\end{align*}
Therefore, the above recovers $\mathcal{J}_{\text{TIS-DPO}}$, leading us to say: 
\begin{equation*}
    \boxed{\mathcal{J}_{\text{MI-DPO}}\left(\pi_{\text{LLM}}, \textcolor{red}{\zeta \propto \zeta_{\text{TIS-DPO}}}\right)\rightarrow\mathcal{J}_{\text{TIS-DPO}}(\pi_{\text{LLM}}).}
\end{equation*}

\end{document}